\documentclass{article}
\usepackage{iclr2027_conference,times}

\usepackage{amsmath,amssymb,amsthm,mathtools,bm}
\usepackage{booktabs}
\usepackage{graphicx}
\usepackage{enumitem}
\usepackage{xcolor}
\usepackage{url}
\usepackage[hidelinks]{hyperref}
\hypersetup{pdftitle={Beyond Group Size: The Spectral Geometry of Plackett--Luce Supervision},pdfauthor={Anjan Goswami}}

\newtheorem{theorem}{Theorem}[section]
\newtheorem{lemma}[theorem]{Lemma}
\newtheorem{proposition}[theorem]{Proposition}
\newtheorem{corollary}[theorem]{Corollary}
\newtheorem{definition}[theorem]{Definition}
\theoremstyle{remark}
\newtheorem{remark}[theorem]{Remark}

\newcommand{\R}{\mathbb{R}}
\newcommand{\E}{\mathbb{E}}
\newcommand{\KL}{\operatorname{KL}}
\newcommand{\diag}{\operatorname{diag}}
\newcommand{\tr}{\operatorname{tr}}
\newcommand{\sech}{\operatorname{sech}}
\newcommand{\1}{\mathbf{1}}
\newcommand{\cL}{\mathcal{L}}
\newcommand{\Pperp}{P_{\perp}}
\newcommand{\Klog}{\kappa_{\log}}
\newcommand{\LBT}{L_{\mathrm{BT}}}
\newcommand{\LPL}{L_{\mathrm{PL}}}
\newcommand{\gPL}{\gamma_{\mathrm{PL}}}
\newcommand{\gsym}{\gamma_{\mathrm{sym}}}
\newcommand{\norm}[1]{\lVert #1\rVert}

\title{Beyond Group Size:\\
The Spectral Geometry of\\
Plackett--Luce Supervision}

\author{Anjan Goswami\thanks{Email: \texttt{anjan@smartinfer.com}. ORCID: \href{https://orcid.org/0009-0002-1778-3454}{0009-0002-1778-3454}.} \\
SmartInfer, Inc.}

\iclrfinalcopy

\begin{document}
\maketitle
\lhead{Preprint}

\begin{abstract}
LLM post-training increasingly uses ranked groups of candidate responses rather than isolated pairwise choices. But a complete ranking is not simply a collection of independent pairwise comparisons: under the Plackett--Luce model, candidates are selected sequentially, so a pair contributes information only while both members remain in the ranking.

We show that the Fisher information in a complete ranking has an exact graph representation with two distinct ingredients. The first is classical pairwise discriminability: candidates with similar scores reveal more about their score difference than candidates whose ordering is nearly certain. The second is listwise exposure: even when a pair's own scores stay fixed, its contribution changes with the other candidates because they determine how long the pair remains jointly active. We derive universal bounds on this exposure and use them to relate the exact Plackett--Luce Fisher operator to a much simpler reward-gap graph. The graph's connectivity then controls the relative-score direction that the ranking identifies least well, within a factor determined by group size.

At complete symmetry, this weakest-direction information increases with group size but quickly saturates, even while ranking entropy and total Fisher information continue to grow. More generally, group size and reward range do not determine the weakest direction. Across 9,696 exact instances, our spectral bounds hold without genuine violation. In a controlled example with the same group size and reward range, weakest-direction curvature changes by a factor of 8384 while total Fisher information moves in the opposite direction; nearly trace-matched configurations can still differ by hundreds-fold. These are local information-geometric results, not finite-sample or downstream-training guarantees.
\end{abstract}

\section{Introduction}
\label{sec:introduction}

Modern preference-learning systems increasingly supervise models with ranked
groups rather than isolated pairwise choices. Pairwise methods learn from a
single binary preference \citep{rafailov2024dpo,azar2024ipo,ethayarajh2024kto};
listwise methods consume an ordering over $K$ candidates
\citep{yuan2023rrhf,zhao2023slichf,liu2024lipo,song2024pro}; and grouped
rollout methods score a batch of samples at once, typically with
group-normalized scalar rewards rather than a ranking likelihood
\citep{shao2024deepseekmath,guo2025grpo}.
We study the complete-ranking case that arises naturally in LLM post-training:
a policy generates $K$ candidate responses to the same prompt, an evaluator
(for example a reward model, verifier, or human) scores or ranks them, and the
resulting grouped feedback is used to update the policy. Our question is how
much information one such ranked group carries about the underlying relative
scores. We model the ranking by the Plackett--Luce distribution, and our results
concern that ranking model; they do not by themselves characterize scalar-reward
objectives such as GRPO's.

\paragraph{From scores to ranking geometry.}
Let $r=(r_1,\ldots,r_K)$ be the candidate scores; they determine a distribution
over complete rankings.  A score direction is identifiable to the extent that
perturbing $r$ along it changes that distribution, and Fisher information
measures this local sensitivity.  A ranking can therefore be informative in
aggregate while leaving one relative-score contrast weakly constrained, and
estimating that contrast then needs many more ranking observations.

This matters because a complete ranking is not equivalent to
observing all $\binom{K}{2}$ pairwise comparisons independently.  Under the
Plackett--Luce (PL) model
\citep{luce1959,plackett1975,yellott1977relationship},
the ranking is generated sequentially: once a candidate is selected, it
disappears from all later stages.  Consequently, the information contributed
by a pair has two conceptually different sources.

The first is classical.  A pair is most informative when its two scores are
close, since the comparison probability is then most sensitive to the gap.  The
unit here is the evaluator's resolution scale $\sigma>0$: ranking probabilities
depend on scores only through $r/\sigma$, so $\sigma$ fixes how large a difference
must be before the evaluator reliably resolves it.  Once the gap is large
relative to $\sigma$ the comparison is nearly deterministic and says little
about the gap's size.

The second is listwise.  Fix $a$ and $b$, so their gap $r_a-r_b$ and their
direct pairwise Fisher information are fixed, and move only a third candidate
$c$.  This changes how likely $c$ is to be selected before either of them, and
hence how long $a$ and $b$ remain jointly active.  Their contribution to the
Fisher information of the \emph{complete ranking} changes even though their own
gap does not.  This paper asks how the two effects combine across the full
ranking.

Adding a constant to every score changes no ranking probability, so only
relative scores are identifiable and the Fisher operator acts on
$V_\perp=\{v\in\mathbb{R}^K : v^\top\mathbf{1}=0\}$.  Its smallest eigenvalue $\gamma_{\mathrm{PL}}$ is the relative-score contrast
the supervision constrains least.  Ranking entropy and Fisher trace do not
determine it: entropy measures uncertainty in the observed ranking and trace
measures total local sensitivity, and neither prevents the information from
concentrating in some directions while one stays weak.

\paragraph{Main result.}
That Fisher operator has an exact representation as a weighted graph on the $K$
candidates, in which the weight of edge $(a,b)$ is a product of two positive
numbers:
\[
    w^{\mathrm{PL}}_{ab}
    =
    \underbrace{\Klog(r_a-r_b;\sigma)}
                _{\text{pairwise discriminability}}
    \times
    \underbrace{\alpha_{ab}(r,\sigma)}
                _{\text{listwise exposure}}
\]
They differ in what each may depend on. The first is the classical
Bradley--Terry Fisher kernel, a function of this pair's own gap $r_a-r_b$ alone.
The second we introduce: the exposure coefficient $\alpha_{ab}$ is a function of
the entire score vector $r$, and measures how much opportunity the sequential
ranking process leaves the pair to contribute.

We prove the universal bounds $1/(K-1)\le\alpha_{ab}(r,\sigma)\le1$, and hence
the operator comparison
\[
    \frac{1}{K-1}L_{\mathrm{BT}}(r,\sigma)
    \preceq
    H_\sigma(r;r)
    \preceq
    L_{\mathrm{BT}}(r,\sigma)
\]
where $H_\sigma(r;r)$ is the exact full-ranking PL Fisher operator and
$L_{\mathrm{BT}}$ is the reward-gap Laplacian whose edge weights are
$\Klog(r_a-r_b;\sigma)$.  Taking the smallest eigenvalue on $V_\perp$ turns this
into the same two-sided bound on $\gamma_{\mathrm{PL}}$ in terms of
$\lambda_2(L_{\mathrm{BT}})$.  Thus the weakest identifiable direction of the full ranking is controlled,
within a factor $K-1$, by the algebraic connectivity of a much simpler
reward-gap graph.  A weak cut in this graph corresponds to a coordinated
relative-score perturbation that changes the ranking distribution only
slightly.

\paragraph{Symmetry as a calibration.}
Complete symmetry, $r_1=\cdots=r_K$, removes score geometry and isolates the
effect of group size $K$.  The Fisher operator is then exactly computable
(Theorem~\ref{thm:symmetric}), and increasing $K$ improves the weakest
identifiable curvature only up to a finite limit, even while ranking entropy
and total Fisher trace continue to grow.  Symmetry is therefore a calibration
rather than the general explanation: once the scores are nonsymmetric, the
geometry of the group matters.  Moreover, positive curvature from individual PL stages
cannot simply be added, because every proper-subset stage is blind to some
directions in the full relative-score space.  This obstruction is why the
operators must be aggregated before their weakest direction is assessed.

\paragraph{Why the weak direction matters.}
For LLM post-training, two groups with the same number of candidate rollouts can
therefore provide very different supervision geometry even before the policy
optimizer or model parameterization enters. Group size and reward range do not
determine the least-constrained score direction, and our experiments show that
even nearly matched total Fisher mass can hide large differences. The gap is
paid in labels: once the weak direction is adequately sampled, estimation error
along it tracks $1/(N\gamma_{\mathrm{PL}})$, so a curvature deficit becomes a
proportional error deficit at equal labelling budget
(Section~\ref{sec:experiments}). We claim no nonasymptotic sample-complexity
theorem or downstream language-model improvement, and in the severely undersampled
regime errors are typically severe rather than gracefully degrading.

\paragraph{How we establish and validate the result.}
We first derive the exact Fisher contribution of each PL stage and show that
every pair separates into two factors: its classical Bradley--Terry
discriminability $\Klog$ and its listwise exposure $\alpha_{ab}$.  A first-hit
argument bounds the cumulative exposure of every pair, and these pairwise
bounds combine to give the spectral sandwich.  We then verify the theory
computationally without reusing the same central
calculation.  For $9{,}696$ score configurations, one implementation computes the
PL Fisher operator directly by enumerating all rankings, while a second
independently reconstructs it from the discriminability--exposure
factorization.  The two agree to numerical precision, and every instance
satisfies the predicted operator bounds.

The novelty is not the Bradley--Terry Fisher kernel, stagewise covariance
decomposition or its equal-score value \citep{hajek2014minimax}, or the general use of graph Laplacians in ranking; it is the
exact arbitrary-score discriminability--exposure representation of the full-ranking PL Fisher
operator, the resulting uniform operator comparison, and its characterization of
the weakest identifiable score direction.  The theoretical claims remain local
score-space Fisher results: we establish no general finite-sample rate and no
downstream language-model guarantee (Section~\ref{sec:discussion} states the
limits in full).

\section{Setup: Ranking Laws and Fisher Geometry}
\label{sec:setup}

For one prompt with $K$ candidates, let $r,u\in\R^K$ be target and model scores, let $\omega\in S_K$ be a complete ranking, and let $S_i(\omega)=\{\omega(i),\dots,\omega(K)\}$ be the remaining set at stage $i$. For evaluator resolution $\sigma>0$ the Plackett--Luce law is
\begin{equation}\label{eq:pl}
P_u^{(\sigma)}(\omega)
=
\prod_{i=1}^{K-1}
\frac{\exp(u_{\omega(i)}/\sigma)}
{\sum_{j\in S_i(\omega)}\exp(u_j/\sigma)}.
\end{equation}
Throughout, the evaluator resolution $\sigma>0$ is fixed and known, and PL probabilities depend on $u$ only through the differences $(u_a-u_b)/\sigma$, so with $\sigma$ fixed the scores are identifiable exactly up to a common additive shift. Adding a constant to all scores changes no ranking probability, so the identifiable relative-score space and its projector are
\[
V_\perp=\{v\in\R^K:v^\top\1=0\},
\qquad
\Pperp=I_K-\tfrac1K\1\1^\top.
\]
Writing $\cL_\sigma(u;r)=\E_{\omega\sim P_r^{(\sigma)}}[-\log P_u^{(\sigma)}(\omega)]$ for the population loss and $H_\sigma(u;r):=\nabla_u^2\cL_\sigma(u;r)$ for its Hessian, we have the exact identity $\cL_\sigma(u;r)-\cL_\sigma(r;r)=\KL(P_r^{(\sigma)}\|P_u^{(\sigma)})$, so at the truth
\begin{equation}\label{eq:fisher}
H_\sigma(r;r)=\nabla_u^2\cL_\sigma(u;r)\big|_{u=r}
\end{equation}
is the Fisher information matrix, and the quantity this paper studies is its weakest identifiable eigenvalue
\[
\gPL(r,\sigma)
=
\min_{\substack{v\in V_\perp\\\norm{v}_2=1}}
v^\top H_\sigma(r;r)v.
\]
We also write $\gPL=\lambda_2(H_\sigma)$, its smallest nonzero eigenvalue.

Throughout, ``information'' means Fisher information about relative-score parameters, especially this weakest identifiable eigenvalue --- not Shannon entropy, mutual information under a prior, or Fisher trace, none of which we claim saturates with $K$.

\paragraph{Two graphs that must not be identified.}
Each PL stage, with active (not yet ranked) candidate set $S$ and selection probabilities $p$, contributes a categorical covariance $\Sigma(p)=\diag(p)-pp^\top$, which decomposes exactly into pairwise contrasts $\sum_{a<b\in S}p_ap_b B_{ab}$ with $B_{ab}:=(e_a-e_b)(e_a-e_b)^\top$ for standard basis vectors $e_a$. Embedding stages into $\R^K$, summing, and averaging over rankings gives the \emph{exact} PL Fisher Laplacian $\LPL(u;r,\sigma)$ with
\[
H_\sigma(u;r)=\LPL(u;r,\sigma),
\]
an identity, not an approximation (Appendix~\ref{app:setupderiv}). Separately, a single direct two-way logistic comparison with gap $\Delta$ has Fisher information
\begin{equation}\label{eq:klog-lbt}
\Klog(\Delta;\sigma)
=
\frac1{4\sigma^2}\sech^2\!\left(\frac{\Delta}{2\sigma}\right),
\qquad
\LBT(r,\sigma)
=
\sum_{a<b}\Klog(r_a-r_b;\sigma)\,B_{ab}.
\end{equation}
These are different objects. $\LPL$ is the exact PL Fisher/Hessian operator~\eqref{eq:fisher}; $\LBT$ is a simpler comparison graph whose weights depend only on target-score gaps relative to $\sigma$; those weights peak at $\Delta=0$ and decay like $e^{-|\Delta|/\sigma}$, so nearly tied candidates form the most informative edges. The exact PL edge coefficient at a stage is $\sigma^{-2}p_a^{(i)}p_b^{(i)}$, \emph{not} $\Klog(r_a-r_b;\sigma)$. Their relation is the theorem of Section~\ref{sec:spectral}, not a definition.

\section{Symmetry: Group Size Helps, but Only Boundedly}
\label{sec:symmetry}

The equal-score Fisher calculation below also appears, at unit resolution, in the Cram{\'e}r--Rao analysis of \citet{hajek2014minimax} (proof of their Theorem~2); we state it in temperature-scaled notation as the symmetric baseline for the arbitrary-score analysis that follows.

\begin{theorem}[Exact symmetric PL curvature]
\label{thm:symmetric}
For $K\ge2$, $\sigma>0$, and $r_1=\cdots=r_K$,
\begin{equation}\label{eq:hsym}
H_{\mathrm{sym},\sigma}
=
\frac1{\sigma^2}\frac{K-H_K}{K-1}\Pperp,
\qquad
H_K=\sum_{m=1}^{K}\frac1m.
\end{equation}
Consequently
\[
\gsym(K,\sigma)
=
\frac1{\sigma^2}\frac{K-H_K}{K-1}.
\]
\end{theorem}

Hence $\gsym(2,\sigma)=1/(2\sigma^2)$, and $\gsym(K,\sigma)$ is strictly increasing in $K$ with
\[
\gsym(K,\sigma)\uparrow\frac1{\sigma^2}
\qquad\text{as }K\to\infty.
\]
Larger symmetric lists therefore do increase the weakest identifiable Fisher curvature; the gain is bounded. Relative to pairwise supervision,
\[
\frac{\gsym(K,\sigma)}{\gsym(2,\sigma)}
=
2\,\frac{K-H_K}{K-1}
<
2
\qquad\text{for every finite }K,
\]
with the ratio tending to $2$ only in the limit: per complete-ranking observation, the weakest-direction gain over one pairwise observation is less than a factor of two.

The trace behaves differently:
\[
\tr(H_{\mathrm{sym},\sigma})
=
\frac{K-H_K}{\sigma^2}
\sim\frac{K}{\sigma^2}.
\]
So total Fisher mass grows roughly linearly while the \emph{weakest identifiable Fisher curvature} saturates. Ranking entropy is a third quantity again: at symmetry it equals $\log_2(K!)$ bits and grows without bound. There is no contradiction: entropy measures uncertainty of the observed ranking, $\gsym$ local distinguishability of nearby score parameters.

The proof averages an embedded subset projector over uniformly random remaining sets and sums a harmonic series; Appendix~\ref{app:symmetric} gives it in full. In expectation the stage with $n$ remaining candidates contributes $\sigma^{-2}\tfrac{n-1}{n(K-1)}\Pperp$, and $\sum_{n=2}^K\tfrac{n-1}{n}=K-H_K$, the numerator in~\eqref{eq:hsym}. Symmetry calibrates the phenomenon but does not explain nonsymmetric groups; the obvious extension, lower-bounding each stage separately, fails for the reason Section~\ref{sec:obstruction} gives.

\section{Why Positive Stagewise Curvature Does Not Naively Add}
\label{sec:obstruction}

At a PL stage with active set $S\subseteq[K]$, $|S|=n$, the categorical Fisher is strictly positive on the local tangent space
\[
T_S
=
\{v\in\R^K:\operatorname{supp}(v)\subseteq S,\;\1_S^\top v=0\}.
\]
Yet proper-subset stages are degenerate on the global identifiable space. Write $\overline{\Sigma}_S(p)$ for the stage covariance $\Sigma(p)$ of Section~\ref{sec:setup} zero-padded outside $S$.

\begin{theorem}[Proper-subset stage obstruction]
\label{thm:obstruction}
Let $\varnothing\neq S\subsetneq[K]$, and let $p_j>0$ on $S$. Define
\[
A_{S,\sigma}(p)=\frac1{\sigma^2}\overline{\Sigma}_S(p).
\]
Then
\[
\lambda_{\min}\!\left(A_{S,\sigma}(p)\big|_{V_\perp}\right)=0,
\]
and
\[
\dim\left(\ker A_{S,\sigma}(p)\cap V_\perp\right)
=
K-|S|.
\]
\end{theorem}

An explicit witness, with $\1_S$ the indicator vector of $S$, is
\[
v_S^\star
=
\1_S-\frac{|S|}{|S^c|}\1_{S^c}.
\]
It is globally mean-zero but constant on $S$, so
\[
A_{S,\sigma}(p)v_S^\star=0.
\]

Each stage operator is strictly positive definite on its own tangent space $T_{S_i}$, which invites lower-bounding $\lambda_{\min}(A_{S_i,\sigma})$ stage by stage and adding the scalars via superadditivity. That step is valid only when every minimum is taken over the \emph{same} space: the positive local minima live on the spaces $T_{S_i}$, whereas the quantity of interest is a minimum over $V_\perp$, where Theorem~\ref{thm:obstruction} gives exactly $0$. This does not make later stages uninformative. Each nullspace moves with the realized remaining set, so differently oriented degenerate operators can sum to a well-conditioned global operator. The lesson is to aggregate operators first and take a minimum eigenvalue last, which is what Section~\ref{sec:spectral} does. Appendix~\ref{app:obstruction} expands both points.

\section{Beyond Symmetry: Discriminability $\times$ Exposure}
\label{sec:spectral}

We now seek the aggregate representation the obstruction requires. The representation below assigns each pair a single edge weight accumulated over all stages before any eigenvalue is taken, so no stagewise minimum, and hence no proper-subset nullspace, enters the bound. Write $\Delta_{ab}=r_a-r_b$, and recall from Section~\ref{sec:setup} the direct pairwise kernel $\Klog(\Delta_{ab};\sigma)$, which is fixed for the pair across all stages, together with the reward-gap Laplacian $\LBT(r,\sigma)=\sum_{a<b}\Klog(\Delta_{ab};\sigma)B_{ab}$. The results below establish the relation between $\LBT$ and the exact operator $H_\sigma(r;r)=\LPL(r;r,\sigma)$.

At a stage $i$ where $a,b\in S_i$, let
\[
p_a^{(i)}
=
\frac{e^{r_a/\sigma}}{\sum_{j\in S_i}e^{r_j/\sigma}},
\qquad
p_b^{(i)}
=
\frac{e^{r_b/\sigma}}{\sum_{j\in S_i}e^{r_j/\sigma}},
\]
and
\[
q_{ab}^{(i)}=p_a^{(i)}+p_b^{(i)}.
\]

\begin{lemma}[Stagewise discriminability--exposure factorization]
\label{lem:stagefactor}
Whenever $a,b\in S_i$,
\[
\frac1{\sigma^2}p_a^{(i)}p_b^{(i)}
=
\left(q_{ab}^{(i)}\right)^2
\Klog(\Delta_{ab};\sigma).
\]
\end{lemma}
The within-pair choice probability $p_a^{(i)}/q_{ab}^{(i)}$ is the stage-independent logistic probability of $\Delta_{ab}/\sigma$, whose Bernoulli variance divided by $\sigma^2$ is $\Klog$ (Appendix~\ref{app:spectral}).

A pair contributes only while both of its members remain active. Define the stage hazard by the piecewise rule
\[
\widetilde q_i^{ab}
:=
\begin{cases}
p_a^{(i)}+p_b^{(i)}, & a,b\in S_i,\\[3pt]
0, & \text{otherwise},
\end{cases}
\qquad
0\le\widetilde q_i^{ab}\le1.
\]
Setting it to zero after either member leaves means it never refers to an inactive candidate's probability. Conditional on the ranking history, $\widetilde q_i^{ab}$ is exactly the hazard that the first member of the pair is selected at stage $i$; Appendix~\ref{app:spectral} gives the filtration-level statement.

\begin{definition}[Listwise exposure coefficient]
\begin{equation}\label{eq:alpha}
\alpha_{ab}(r,\sigma)
=
\E\left[
\sum_{i=1}^{K-1}
(\widetilde q_i^{ab})^2
\right].
\end{equation}
\end{definition}

The square is inherited from Lemma~\ref{lem:stagefactor}, where a stage contributes the pair's joint mass squared.

\begin{theorem}[Exact exposure-weighted Fisher representation]
\label{thm:factorization}
For every finite $r\in\R^K$ and $\sigma>0$,
\begin{equation}\label{eq:factorization}
H_\sigma(r;r)
=
\sum_{a<b}
\alpha_{ab}(r,\sigma)
\Klog(r_a-r_b;\sigma)
(e_a-e_b)(e_a-e_b)^\top.
\end{equation}
\end{theorem}
Its proof averages the exact stage covariances $\sigma^{-2}\sum p_a^{(i)}p_b^{(i)}B_{ab}$ over rankings and applies Lemma~\ref{lem:stagefactor} at every stage where both $a$ and $b$ are active (Appendix~\ref{app:spectral}).

Thus
\[
\boxed{
\text{effective edge information}
=
\text{pair discriminability}
\times
\text{listwise exposure}.
}
\]

Let $\tau_{ab}=\min\{i:\omega(i)\in\{a,b\}\}$ be the stage at which the first member of the pair is selected. The events $\{\tau_{ab}=i\}$, $i=1,\dots,K-1$, are disjoint and exhaustive, and summing the stagewise hazards gives
\begin{equation}\label{eq:firsthit}
\E\left[\sum_{i=1}^{K-1}\widetilde q_i^{ab}\right]=1.
\end{equation}
The identity holds in expectation, not pathwise: $\sum_i\widetilde q_i^{ab}$ need \emph{not} equal $1$ on a realized ranking, and the lower bound below is proved so as not to require it.

\begin{theorem}[Universal exposure bounds]
\label{thm:alphabounds}
For every pair, finite $r$, and $\sigma>0$,
\[
\frac1{K-1}\le\alpha_{ab}(r,\sigma)\le1.
\]
\end{theorem}
The upper bound uses $(\widetilde q_i^{ab})^2\le\widetilde q_i^{ab}$ and the first-hit identity~\eqref{eq:firsthit}; the lower bound applies Cauchy--Schwarz pathwise over the at most $K-1$ stages, $\sum_i(\widetilde q_i^{ab})^2\ge Z^2/(K-1)$ with $Z=\sum_i\widetilde q_i^{ab}$, and then Jensen, $\E[Z^2]\ge(\E Z)^2=1$ (Lemma~\ref{lem:firsthit}).

Because each $B_{ab}$ is positive semidefinite, these edgewise bounds lift to the reward-gap Laplacian $\LBT(r,\sigma)$ of Section~\ref{sec:setup} (multiply by $\Klog B_{ab}$, sum over pairs, and use Theorem~\ref{thm:factorization}; Appendix~\ref{app:spectral}).

\begin{theorem}[Spectral sandwich]
\label{thm:sandwich}
For every finite $r\in\R^K$ and $\sigma>0$,
\begin{equation}\label{eq:sandwich}
\frac1{K-1}\LBT(r,\sigma)
\preceq
H_\sigma(r;r)
\preceq
\LBT(r,\sigma).
\end{equation}
Consequently,
\[
\frac1{K-1}\lambda_2(\LBT(r,\sigma))
\le
\gPL(r,\sigma)
\le
\lambda_2(\LBT(r,\sigma)).
\]
\end{theorem}

Here $\preceq$ is the Loewner order ($B-A$ positive semidefinite), so the comparison holds in every direction at once, and $\lambda_2$ is the graph's algebraic connectivity. This is a factor-$(K-1)$ characterization of exact PL Fisher curvature by a graph whose weights depend only on reward gaps relative to evaluator resolution. Geometrically $\LBT$ is heavy between candidates with close scores, so scores clumped into well-separated groups leave exponentially light edges between the groups and nearly cut the graph; the weakest direction is that between-group contrast.

\paragraph{Boundary checks.}
Three cases pin the normalization down.
For $K=2$ there is a single stage with $\widetilde q_1^{12}=1$, so $\alpha_{12}=1$ and
\[
H_\sigma(r;r)=\Klog(r_1-r_2;\sigma)B_{12}
\]
is exactly the Bradley--Terry Fisher matrix; the sandwich collapses to equality because $K-1=1$.
For $K=3$ with first-stage pair mass $q=p_a^{(1)}+p_b^{(1)}$, either the pair is hit immediately or the third candidate goes first and the pair mass becomes $1$, so
\begin{equation}\label{eq:k3alpha}
\alpha_{ab}=q^2+(1-q)=1-q+q^2=\left(q-\tfrac12\right)^2+\tfrac34 \ge \tfrac34,
\end{equation}
which is strictly stronger than the $1/2$ guaranteed by Theorem~\ref{thm:alphabounds}. Because $q$ changes when the third candidate moves even though $r_a-r_b$ is fixed, this is the smallest explicit instance of pure listwise exposure; the universal constant is deliberately conservative, and whether $1/(K-1)$ can be improved for general $K$ is open. For $K=3$ itself, \eqref{eq:k3alpha} and~\eqref{eq:factorization} give $\tfrac34\LBT\preceq H_\sigma\preceq\LBT$, and $\tfrac34$ is sharp: for $r=(0,0,-t)$ and $v\propto(-\tfrac12,-\tfrac12,1)$ the edge $\{1,2\}$ drops out and $v^\top H_\sigma v/v^\top\LBT v=1-q+q^2$ with $q=(1+e^{-t/\sigma})/(2+e^{-t/\sigma})\to\tfrac12$ as $t\to\infty$, so the constant is approached but not attained.
At symmetry, summing $\Pr(a,b\in S_n)\,(2/n)^2=4(n-1)/(nK(K-1))$ over $n=2,\dots,K$ gives
\[
\alpha_{ab}^{\mathrm{sym}}
=
\frac{4(K-H_K)}{K(K-1)},
\]
so every effective edge has weight $(K-H_K)/(\sigma^2K(K-1))$, and the complete-graph Laplacian's nonzero eigenvalue is $K$ times that, recovering Theorem~\ref{thm:symmetric}. The saturation $\gsym\uparrow1/\sigma^2$ is an exact symmetric statement from Theorem~\ref{thm:symmetric}, not a consequence of the range-free sandwich~\eqref{eq:sandwich}: at symmetry $\lambda_2(\LBT)=K/(4\sigma^2)$, so the sandwich's upper bound grows linearly in $K$, while its conservative lower bound stays within a factor $4(K-H_K)/K<4$ of $\gsym$.

\begin{corollary}[Group size and reward range do not determine curvature]
\label{cor:separation}
Fix $K=4$ and $R>0$. Let
\[
r^{\mathrm{cluster}}=(-R,-R,R,R),
\qquad
r^{\mathrm{spread}}
=
\left(-R,-\frac R3,\frac R3,R\right).
\]
Both have range $2R$, but
\[
\frac{\gPL(r^{\mathrm{spread}},\sigma)}
{\gPL(r^{\mathrm{cluster}},\sigma)}
\ge
\frac{2-\sqrt2}{12}
\frac{\sech^2(R/(3\sigma))}
{\sech^2(R/\sigma)}.
\]
Hence the ratio grows as $\Omega(\exp(4R/(3\sigma)))$ as $R/\sigma\to\infty$.
\end{corollary}

\begin{corollary}[Greedy group selection]
\label{cor:design}
Fix $K$ and $\sigma$ and let $\mathcal{S}$ be a finite collection of candidate groups of size $K$. If $S_E$ maximizes $\lambda_2(\LBT(S,\sigma))$ over $\mathcal{S}$ and $S^\star$ maximizes $\gPL(S,\sigma)$ over $\mathcal{S}$, then
\begin{equation}\label{eq:greedy}
\gPL(S_E,\sigma) \ge \tfrac{1}{K-1}\,\gPL(S^\star,\sigma).
\end{equation}
\end{corollary}

This is immediate from Theorem~\ref{thm:sandwich}, chaining its lower bound at $S_E$, the maximality of $\lambda_2$, and its upper bound at $S^\star$. It says that selecting groups on the computable $\lambda_2(\LBT)$ costs at most a factor $K-1$ against selecting on the exact curvature; designing and evaluating acquisition policies is outside the scope of this paper. Operationally, for a finite family of equal-size candidate groups with target or estimated scores, one forms $\LBT$ for each group from its score gaps via~\eqref{eq:klog-lbt}, without collecting any ranking, and keeps the group with the largest $\lambda_2$. With estimated scores the guarantee becomes $e^{-2\delta/\sigma}/(K-1)$ when every within-group score-gap error is at most $\delta$ (Appendix~\ref{app:rebuttal:estimated}); Appendix~\ref{app:separation} gives an exact two-group illustration. The factor-$(K-1)$ guarantee is worst-case; we do not characterize its tightness over realistic candidate pools here.

\section{What Simple Summaries Miss}
\label{sec:experiments}

The theory shows that group size and reward range do not determine the weakest identifiable Fisher curvature. Our controlled deformation makes the distinction quantitative: with both held fixed, $\gPL$ changes by $8384\times$ while total Fisher trace changes by about $35\times$ in the opposite direction. Practically, selection or sample-allocation rules based only on these summaries can therefore treat preference groups with substantially different information geometry as equivalent. We then ask whether the trace itself can nevertheless serve as a sufficient summary. We pre-specified $9{,}696$ exact PL instances with $K\in\{3,4,5,6\}$, $\sigma=1$, six score scales, and eight families, exhaustively enumerating all $K!$ rankings. Because $\Klog$ depends on scores only through $\Delta/\sigma$, varying score scale at fixed $\sigma$ sweeps the gap-to-resolution regime rather than a single evaluator resolution. We independently validate the Fisher matrices by two code paths: embedded stage covariances and the survival-aware $\alpha_{ab}$ construction of Theorem~\ref{thm:factorization}. Their maximum relative Frobenius discrepancy is $1.03\times10^{-14}$, below the pre-specified $10^{-10}$ gate, with all ten blocking checks passing. Since $\rho=\lambda_2(H_\sigma)/\lambda_2(\LBT)$ is ill-conditioned near the noise floor, PSD comparisons are primary; Appendix~\ref{app:expdetails} gives the full protocol and conditioning analysis.

\subsection{Independent verification of the factorization and spectral sandwich}

\begin{figure}[t]
\centering
\includegraphics[width=4.60in]{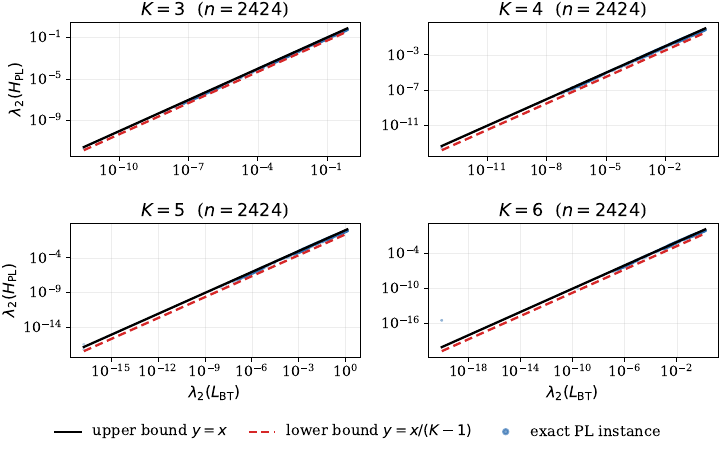}
\caption{Exact PL weakest identifiable Fisher curvature against reward-gap graph connectivity, all $9{,}696$ instances, one panel per $K$. Solid: upper bound $\lambda_2(\LBT)$; dashed: lower bound $\lambda_2(\LBT)/(K-1)$. Every instance lies inside the band of Theorem~\ref{thm:sandwich}.}
\label{fig:envelope}
\end{figure}

Across all instances both PSD inequalities held to roundoff (minimum eigenvalues $-6.1\times10^{-15}$ and $-2.3\times10^{-15}$) (Figure~\ref{fig:envelope}), and the exposure bounds of Theorem~\ref{thm:alphabounds} held for every pair. Thirteen unstable eigenvalue-ratio cases occurred only when both eigenvalues sat at the numerical noise floor; all satisfied the bounds at $50$-digit precision (Appendix~\ref{app:expdetails}).

\subsection{Fixed group size and range do not determine weakest identifiable Fisher curvature}
\label{sec:controlled}

\begin{figure}[t]
\centering
\includegraphics[width=4.60in]{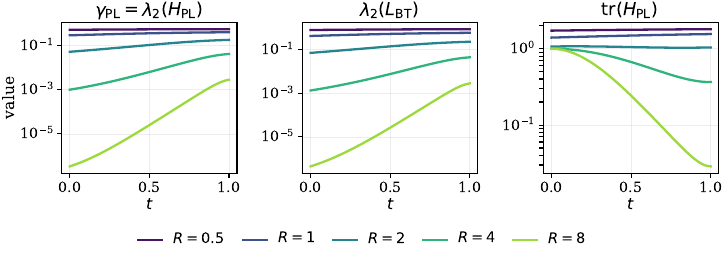}
\caption{Controlled $K=4$ deformation with range held at exactly $2R$, from clustered $(-R,-R,R,R)$ at $t=0$ to spread $(-R,-R/3,R/3,R)$ at $t=1$. Left: $\gPL$; centre: $\lambda_2(\LBT)$; right: $\tr(H_\sigma)$; curves are the five values of $R$.}
\label{fig:path}
\end{figure}

To test Corollary~\ref{cor:separation} without confounds we interpolated between its two configurations at $K=4$, verifying programmatically that the range equals $2R$ at every point to machine precision. The effect is large and grows with $R/\sigma$, as the theory predicts (full sweep in Appendix~\ref{app:expdetails}, Table~\ref{tab:path}). At $R=8$, weakest-direction curvature rises from $\gPL=3.38\times10^{-7}$ at the clustered endpoint to $2.83\times10^{-3}$ at the spread endpoint, a factor of $8384$, while the Fisher trace \emph{falls} by about $35\times$: total Fisher mass and weakest-direction curvature move in opposite directions along this controlled deformation, and neither $K$ nor the range changes anywhere along it. That conditioning has a measurable label cost. In the adequately sampled regime $N\gPL\ge10$, identified post hoc, weak-direction estimation error follows the expected $1/(N\gPL)$ behaviour, with log-log slope $-1.0045$ ($R^2=0.996$, no free parameters). At $R=2$, well inside this regime, the two configurations differ by $3.46\times$ in weakest-direction curvature and show a comparably sized difference in weak-direction error at fixed $N$. Below roughly ten effective observations in that direction most cells show severe estimation errors rather than graceful degradation, so the frozen form of this hypothesis, which set no such regime restriction, is \emph{not} supported as stated and the restriction is post hoc (Appendix~\ref{app:labelcost}).

\subsection{Nearly matched Fisher trace can hide large curvature differences}

Section~\ref{sec:controlled} holds $K$ and the range fixed but lets the trace move freely --- indeed opposite to $\gPL$. This experiment is its adversarial extension, asking whether the separation survives once the trace is \emph{also} approximately matched, which closes the escape route that total Fisher mass explains the effect. Theorem~\ref{thm:sandwich} says nothing about the trace, so this is empirical. We ran a pre-specified search over $7{,}424$ candidates per $(K,\text{range})$ cell, scanning exhaustively within each trace-tolerance band so the reported optimum is the bank optimum, not a hand-picked example.

At matched $K$ and matched range, we find realizable score configurations whose Fisher traces agree to within $1\%$ but whose weakest identifiable Fisher curvatures differ by hundreds-fold. At range $16$ and $1\%$ trace tolerance the best curvature ratios were $127.5\times$ at $K=4$ (traces agreeing to $0.71\%$), $249.5\times$ at $K=5$ ($0.37\%$), and $478.3\times$ at $K=6$ ($0.68\%$). The tolerance ladder never had to be relaxed; the $1\%$ band already yields these separations. Score vectors for all three pairs are in Appendix~\ref{app:expdetails}, Table~\ref{tab:tracematched}, together with the trace/curvature scatter (Figure~\ref{fig:trace}); both members of every pair were reverified at $50$ decimal digits.

This is an \emph{empirical separation over the tested candidate families, not a theorem}: it shows such configurations exist and are easy to find, which rules out total Fisher mass as a sufficient summary, but establishes no general statement about the trace.

Correlation between $\lambda_2(H_\sigma)$ and $\lambda_2(\LBT)$ is not independent evidence here: Theorem~\ref{thm:sandwich} already confines the ratio to $[1/(K-1),1]$, so a high correlation is forced and could not have come out otherwise. Measuring it checks the implementation and how tight the envelope is in practice, not whether $\lambda_2(\LBT)$ is predictive. Appendix~\ref{app:expdetails} gives the descriptive comparison, the within-$K$ breakdown, and the realized ratios.

\section{Related Work}
\label{sec:related_work}

\paragraph{PL inference, comparison topology, and spectral structure.}
Bradley--Terry and Plackett--Luce are classical \citep{bradley1952rank,luce1959,plackett1975,hunter2004mm,yellott1977relationship}, and a literature ties ranking estimation to graph structure: rates governed by algebraic connectivity of item-pair co-occurrence \citep{vojnovic2023accelerated}, structure-dependent risk bounds for repeated selection \citep{seshadri2020rich}, minimax rates whose Cram{\'e}r--Rao bound depends inversely on the spectral gap of a comparison-graph Laplacian \citep{hajek2014minimax}, and unified asymptotics under deterministic comparison-graph conditions \citep{hanxu2025unified}. Closest to our setting, \citet{hajek2014minimax} study PL rankings of assigned candidate subsets: their Cram{\'e}r--Rao analysis writes the Fisher matrix through pairwise stage terms and evaluates it exactly at equal scores (the value in Theorem~\ref{thm:symmetric}), and their maximum-likelihood analysis lower-bounds curvature by a score-independent co-occurrence Laplacian with constants that depend on the score range. We instead factor each edge, at arbitrary scores, into Bradley--Terry discriminability and listwise exposure, which gives a range-free comparison with a score-dependent Laplacian and consequences for weakest-direction curvature and group selection. Our design is instead fixed to one complete ranking, whose internally generated Fisher graph we characterize.

\paragraph{Three Laplacians that must be kept distinct.}
\emph{Comparison-design Laplacians} encode which items co-occur across observations, a property of the collection protocol. Our $\LBT(r,\sigma)$ is not one: it is score-dependent, its weights the Fisher sensitivities of hypothetical direct Bradley--Terry comparisons. The third, $H_\sigma(r;r)=\LPL(r;r,\sigma)$, is the exact Fisher operator of a complete PL ranking. Theorem~\ref{thm:factorization} connects the latter two edge by edge, and Theorem~\ref{thm:sandwich} turns the exposure bounds into a two-sided operator comparison. Our contribution is this arbitrary-score discriminability--exposure factorization of the full-ranking score-space Fisher operator and the resulting range-free comparison with the reward-gap Laplacian.

\paragraph{Subset-wise feedback, design, and listwise objectives.}
That larger comparison sets need not give proportional gains is known: winner-only subset feedback barely improves on pairwise as the subset grows, while top-$m$ feedback does \citep{saha2019active}, and full ranking feedback can improve regret in online subset selection \citep{saha2019relative}. Those study repeated adaptive interaction; we ask how much local information about the relative-score vector is in \emph{one} complete ranking. A second line asks which comparisons to acquire: D-optimal design for $K$-subsets \citep{thekumparampil2025few}, PL-specific Fisher objectives for negative selection \citep{surana2026massdpo}, and classical optimal design \citep{pukelsheim2006optimal}. Our goal is prior to acquisition, so $\alpha_{ab}$ is not a sampling probability but the attenuation the sequential PL process applies to edge $(a,b)$. Listwise and grouped alignment objectives \citep{yuan2023rrhf,zhao2023slichf,liu2024lipo,song2024pro,shao2024deepseekmath,guo2025grpo} alongside pairwise DPO, IPO and KTO \citep{rafailov2024dpo,azar2024ipo,ethayarajh2024kto} motivate the setting, and recent theory writes the Hessian of a log-linear PL listwise loss as stagewise conditional covariances \citep{wu2026dro_lpo}. Beyond the priority disclaimers of Section~\ref{sec:introduction}, we do \emph{not} claim the nullspace statement behind Theorem~\ref{thm:obstruction} is anything but elementary; the contribution is the exact factorization, the two-sided spectral comparison, and the structural role of the obstruction.

\section{Discussion and Conclusion}
\label{sec:discussion}

A full ranking is governed not only by which pairs are discriminable, but by how the sequential process exposes those pairs and how their information connects across relative-score space. The exact PL Fisher edge weight factors as $w_{ab}^{\mathrm{PL}}=\alpha_{ab}\Klog(\Delta_{ab};\sigma)$, and universal exposure bounds place the operator between $\LBT/(K-1)$ and $\LBT$: grouped preference information is geometric rather than merely combinatorial.

\paragraph{What this gives a practitioner.}
The reward-gap Laplacian is computable from target or estimated scores and the evaluator-resolution scale alone, so $\lambda_2(\LBT)$ is available \emph{before} a ranking is collected. Selecting groups on it costs at most a factor $K-1$ against exact-curvature selection (Corollary~\ref{cor:design}); Section~\ref{sec:experiments} shows that equal-size, equal-range groups can still differ sharply in weakest-direction curvature. That conditioning is paid in labels: well inside the adequately sampled regime, a $3.46\times$ curvature deficit shows up as a comparably sized error deficit at equal budget.

\paragraph{What this does not establish.}
All main theorems characterize the Fisher operator at the truth: no global quadratic growth, no nonasymptotic sample-complexity rate, no convergence rate for a nonlinear policy optimizer, and the experiments show no downstream language-model consequence and no evidence that $\lambda_2(\LBT)$ predicts curvature outside the sampled geometries; Appendix~\ref{app:historical}'s historical results are directional and used a trace-based group utility, so they do not test Corollary~\ref{cor:design}. Conditioning may weaken under $J_\theta^\top H_\sigma J_\theta$; Corollary~\ref{cor:design} motivates investigating spectral group selection in ranking-based preference learning; its downstream effectiveness remains open.

\subsubsection*{Reproducibility Statement}
Complete proofs are provided in the appendices. The frozen experimental specifications, exact-enumeration implementation, independent computational cross-checks, numerical validation gates, and scripts used to reproduce every table and figure generated by the spectral-geometry and label-cost experiments are available from the author on request; the historical results in Appendix~\ref{app:historical} come from earlier, separately run experiments for which no artifacts are provided. The experiment was frozen before result inspection, and numerically degenerate cases were additionally recomputed at high precision.

\subsubsection*{AI Use Statement}
Generative AI tools were used as research assistants during the development of this work. They assisted with brainstorming and refining theoretical formulations, checking mathematical arguments and proofs, designing and implementing computational experiments, including code for generating synthetic score configurations and ranking observations, analyzing and stress-testing results, surveying and organizing related literature, and drafting and editing portions of the manuscript. All AI-assisted mathematical claims, code, and experimental results were independently checked against explicit derivations, validation gates, or frozen artifacts. The author takes responsibility for the final content of the work.

\bibliographystyle{iclr2027_conference}
\bibliography{iclr_information_geometry_v0}

\appendix

\section{Setup Derivations}
\label{app:setupderiv}

This appendix records the derivations compressed out of Section~\ref{sec:setup}. Nothing here is new; it is the explicit form of the identities quoted there.

\paragraph{Pairwise decomposition of a categorical Fisher.}
For a categorical probability vector $p$ supported on an active set $S$,
\[
\Sigma(p)=\diag(p)-pp^\top
=
\sum_{a<b\in S}
p_ap_b(e_a-e_b)(e_a-e_b)^\top .
\]
Thus a PL stage is already a weighted graph of pairwise contrast directions, with edge weight $p_ap_b$ on the pair $(a,b)$.

\paragraph{The exact PL Fisher operator is a graph Laplacian.}
Embedding each stage covariance into the original $K$ coordinates, summing stages, and averaging over rankings defines
\[
\LPL(u;r,\sigma)
:=
\frac1{\sigma^2}
\sum_{i=1}^{K-1}
\E_{\omega\sim P_r^{(\sigma)}}
\left[
\sum_{a<b\in S_i(\omega)}
p_a^{(i)}(u,\omega)\,p_b^{(i)}(u,\omega)\,
B_{ab}
\right],
\qquad
B_{ab}:=(e_a-e_b)(e_a-e_b)^\top,
\]
and by construction $H_\sigma(u;r)=\LPL(u;r,\sigma)$. This is an identity rather than an approximation. Its edge weights depend on the stage, the realized active set, and the full PL probabilities, which is precisely why the exact operator is not the reward-gap graph $\LBT$.

\paragraph{Equivalent form of the pairwise kernel.}
The direct two-way logistic Fisher kernel admits the equivalent exponential form
\[
\Klog(\Delta;\sigma)
=
\frac1{4\sigma^2}\sech^2\!\left(\frac{\Delta}{2\sigma}\right)
=
\frac1{\sigma^2}
\frac{e^{r_a/\sigma}e^{r_b/\sigma}}{\left(e^{r_a/\sigma}+e^{r_b/\sigma}\right)^2},
\qquad \Delta=r_a-r_b,
\]
which is the form used in the numerically stable implementation of Section~\ref{sec:experiments}.

\section{Proof of the Symmetric Curvature Theorem}
\label{app:symmetric}

For $S\subseteq[K]$ with $|S|=n$, define
\[
P_{S,\perp}^{\rm emb}
=
D_S-\frac1n\1_S\1_S^\top.
\]
At symmetry the stage Fisher for remaining set $S$ is
\[
\frac1{\sigma^2n}P_{S,\perp}^{\rm emb}.
\]

\paragraph{Proof mechanics in brief.}
At a stage with $n$ remaining candidates, the local symmetric Fisher is $\sigma^{-2}n^{-1}$ times the local mean-zero projector. Averaging that projector over all uniformly random size-$n$ remaining sets yields
\[
\E[P_{S,\perp}^{\mathrm{emb}}]
=
\frac{n-1}{K-1}\Pperp,
\]
and summing over $n=K,\dots,2$ gives
\[
\sum_{n=2}^K\frac{n-1}{n}=K-H_K.
\]

\begin{lemma}[Uniform remaining set]
At the stage with $n$ remaining candidates, $S$ is uniform over the $\binom Kn$ size-$n$ subsets.
\end{lemma}

\begin{proof}
Fix $S$ with $|S|=n$. The $K-n$ removed items can occupy the first $K-n$ positions in $(K-n)!$ orders and the items in $S$ can occupy the last $n$ positions in $n!$ orders. Thus the event has probability $(K-n)!n!/K!=\binom Kn^{-1}$.
\end{proof}

\begin{lemma}[Average subset projector]
\[
\E_S[P_{S,\perp}^{\rm emb}]
=
\frac{n-1}{K-1}\Pperp.
\]
\end{lemma}

\begin{proof}
Permutation invariance implies
\[
M_n:=\E[P_{S,\perp}^{\rm emb}]
=
aI+b\1\1^\top.
\]
Every subset projector annihilates $\1$, so $M_n=a\Pperp$. Taking traces,
\[
n-1=\tr(M_n)=a(K-1),
\]
which gives $a=(n-1)/(K-1)$.
\end{proof}

\begin{proof}[Proof of Theorem~\ref{thm:symmetric}]
Summing expected stage contributions over $n=K,\dots,2$,
\[
H_{\rm sym,\sigma}
=
\frac1{\sigma^2(K-1)}
\sum_{n=2}^K\frac{n-1}{n}\Pperp.
\]
Since
\[
\sum_{n=2}^K\frac{n-1}{n}=K-H_K,
\]
the result follows.
\end{proof}

\section{Proof of the Stagewise Obstruction}
\label{app:obstruction}

\begin{proof}[Proof of Theorem~\ref{thm:obstruction}]
Let $n=|S|$ and $m=|S^c|=K-n$. Define
\[
v_S^\star=\1_S-\frac nm\1_{S^c}.
\]
Then $(v_S^\star)^\top\1=0$, so $v_S^\star\in V_\perp$. On $S$, the vector is constant. Because
\[
\Sigma_S(p)\1_S=0,
\]
the embedded Fisher satisfies
\[
A_{S,\sigma}(p)v_S^\star=0.
\]
Since $A_{S,\sigma}(p)$ is PSD, its restricted minimum Rayleigh quotient on $V_\perp$ is zero.

For the nullity statement, positivity of every $p_j$ implies the local categorical Fisher has nullspace exactly the constant direction on $S$. After embedding, coordinates on $S^c$ are arbitrary, so
\[
\dim\ker A_{S,\sigma}=1+(K-n).
\]
Intersecting with $v^\top\1=0$ leaves dimension $K-n$.
\end{proof}

\paragraph{The invalid argument this rules out.}
Each stage operator is strictly positive definite on its own tangent space $T_{S_i}$, which invites the following reasoning: lower-bound $\lambda_{\min}(A_{S_i,\sigma})$ stage by stage, then add the scalars using superadditivity,
\[
\lambda_{\min}\!\left(\sum_i A_{S_i,\sigma}\right)
\ge
\sum_i\lambda_{\min}(A_{S_i,\sigma}).
\]
The superadditivity step is valid only when every minimum is taken over the same space. Here the positive local minima live on the spaces $T_{S_i}$, whereas the quantity of interest is a minimum over $V_\perp$, and Theorem~\ref{thm:obstruction} shows that each proper-subset stage contributes exactly $0$ there. The scalars being added are curvatures of different spaces, so the conclusion does not follow.

\paragraph{What remains true.}
The theorem does not say that later stages are uninformative. Each proper-subset stage is positive on its own local tangent space and globally degenerate on $V_\perp$, and its nullspace moves with the realized remaining set. Sums and expectations of differently oriented degenerate operators can therefore recover positive global curvature. The lesson is to aggregate operators first and take a minimum eigenvalue last, which is what Section~\ref{sec:spectral} does.

\section{Proofs for Exposure Factorization and Spectral Sandwich}
\label{app:spectral}

\begin{proof}[Proof of Lemma~\ref{lem:stagefactor}]
When $a,b\in S_i$,
\[
\frac{p_a^{(i)}}{p_a^{(i)}+p_b^{(i)}}
=
\frac{e^{r_a/\sigma}}{e^{r_a/\sigma}+e^{r_b/\sigma}}.
\]
Therefore
\[
\frac1{\sigma^2}p_a^{(i)}p_b^{(i)}
=
\left(p_a^{(i)}+p_b^{(i)}\right)^2
\Klog(r_a-r_b;\sigma).
\]
\end{proof}

\begin{proof}[Proof of Theorem~\ref{thm:factorization}]
For one fixed pair, with $X_i^{ab}=\1\{a,b\in S_i\}$ the indicator that both $a$ and $b$ remain active at stage $i=1,\dots,K-1$,
\[
w_{ab}^{\rm PL}
=
\E\left[
\sum_i
X_i^{ab}
\frac1{\sigma^2}p_a^{(i)}p_b^{(i)}
\right].
\]
Applying Lemma~\ref{lem:stagefactor},
\[
w_{ab}^{\rm PL}
=
\E\left[
\sum_i
(\widetilde q_i^{ab})^2
\right]
\Klog(\Delta_{ab};\sigma)
=
\alpha_{ab}\Klog(\Delta_{ab};\sigma).
\]
Summing all pair edges gives the theorem.
\end{proof}

\begin{lemma}[First-hit compensator identity]
\label{lem:firsthit}
For every pair,
\[
\E\left[\sum_{i=1}^{K-1}\widetilde q_i^{ab}\right]=1.
\]
\end{lemma}

\begin{proof}
Let $\mathcal{F}_i=\sigma(\omega(1),\dots,\omega(i-1))$ and let $\tau_{ab}=\min\{i:\omega(i)\in\{a,b\}\}$. The event that both $a$ and $b$ are still active at stage $i$ is $\mathcal{F}_i$-measurable, and by the piecewise definition $\widetilde q_i^{ab}$ is $\mathcal{F}_i$-measurable and never evaluates a probability of an inactive candidate. Conditioned on $\mathcal{F}_i$, stage $i$ selects the first member of the pair with probability exactly $\widetilde q_i^{ab}$, so
\[
\Pr(\tau_{ab}=i\mid\mathcal{F}_i)=\widetilde q_i^{ab}.
\]
Taking total expectation gives the pointwise-in-$i$ identity $\E[\widetilde q_i^{ab}]=\Pr(\tau_{ab}=i)$. The events $\{\tau_{ab}=i\}$ for $i=1,\dots,K-1$ are disjoint and exhaustive: a complete ranking eventually selects one of $a,b$, and if the two are the final survivors this happens at stage $K-1$. Summing over $i$ gives the claim.
\end{proof}

\begin{remark}
The stronger pathwise statement $\sum_i\widetilde q_i^{ab}=1$ is false in general and is not used. Only the expectation is pinned to $1$, which is why the lower bound below applies Cauchy--Schwarz pathwise and passes to expectations afterwards.
\end{remark}

\begin{proof}[Proof of Theorem~\ref{thm:alphabounds}]
\emph{Upper bound.} Since $0\le\widetilde q_i^{ab}\le1$ we have $(\widetilde q_i^{ab})^2\le\widetilde q_i^{ab}$ pathwise, so by Lemma~\ref{lem:firsthit}
\[
\alpha_{ab}
=
\E\sum_i(\widetilde q_i^{ab})^2
\le
\E\sum_i\widetilde q_i^{ab}
=
1.
\]

\emph{Lower bound.} There are at most $K-1$ nontrivial stages, so on every realized ranking path Cauchy--Schwarz gives
\[
\sum_{i=1}^{K-1}(\widetilde q_i^{ab})^2
\ge
\frac1{K-1}\left(\sum_{i=1}^{K-1}\widetilde q_i^{ab}\right)^2.
\]
Taking expectations and then applying Jensen's inequality to $Z=\sum_i\widetilde q_i^{ab}$,
\[
\alpha_{ab}
\ge
\frac1{K-1}\E[Z^2]
\ge
\frac1{K-1}(\E Z)^2
=
\frac1{K-1},
\]
the last equality by Lemma~\ref{lem:firsthit}. Note the order: the pathwise inequality comes first and the expectation identity is applied only at the end, so no pathwise normalization of $\sum_i\widetilde q_i^{ab}$ is assumed.
\end{proof}

\begin{proof}[Proof of Theorem~\ref{thm:sandwich}]
Every edge matrix $B_{ab}=(e_a-e_b)(e_a-e_b)^\top$ is positive semidefinite, so Theorem~\ref{thm:alphabounds} gives the edgewise Loewner inequalities
\[
\frac1{K-1}\Klog(\Delta_{ab};\sigma)B_{ab}
\preceq
\alpha_{ab}\Klog(\Delta_{ab};\sigma)B_{ab}
\preceq
\Klog(\Delta_{ab};\sigma)B_{ab}.
\]
Summing over pairs preserves the Loewner order, and Theorem~\ref{thm:factorization} identifies the middle sum as $H_\sigma(r;r)$, which gives the operator sandwich. All three operators are graph Laplacians and therefore annihilate $\1$, so restricting the common Rayleigh quotient to
\[
V_\perp=\{v:v^\top\1=0\}
\]
preserves the order of the minima. The minimum of the Rayleigh quotient of a Laplacian over unit vectors in $V_\perp$ is its second-smallest eigenvalue, which yields
\[
\frac1{K-1}\lambda_2(\LBT)
\le
\gPL(r,\sigma)
\le
\lambda_2(\LBT).
\]
No commuting or shared eigenvectors are required.
\end{proof}

\section{Proof of the Fixed-Range Separation}
\label{app:separation}

For
\[
r^{\rm cluster}=(-R,-R,R,R),
\]
writing $w(\Delta):=\Klog(\Delta;\sigma)$ for the kernel of Section~\ref{sec:setup}, the two within-cluster BT edges have weight $w(0)$ and all four cross-cluster edges have weight
\[
w(2R).
\]
The nonzero eigenvalues of $\LBT(r^{\rm cluster},\sigma)$ are
\[
4w(2R),
\qquad
2\bigl(w(0)+w(2R)\bigr),
\qquad
2\bigl(w(0)+w(2R)\bigr),
\]
the first belonging to the cluster-contrast vector $(1,1,-1,-1)^\top$ and the other two to the within-cluster contrasts. Since $w(0)\ge w(2R)$, the smallest of these is the first. Hence
\[
\lambda_2(\LBT(r^{\rm cluster},\sigma))
=
4w(2R),
\]
so
\[
\gPL(r^{\rm cluster},\sigma)\le4w(2R).
\]

For
\[
r^{\rm spread}
=
\left(-R,-\frac R3,\frac R3,R\right),
\]
the three adjacent score gaps all equal $2R/3$, so the BT graph contains, as a subgraph, the four-vertex path $1-2-3-4$ with all three edge weights $w(2R/3)$. Deleting the remaining edges removes positive semidefinite terms and therefore only decreases the Laplacian in Loewner order. The algebraic connectivity of the unweighted four-vertex path is $2-\sqrt2$, so
\[
\lambda_2(\LBT(r^{\rm spread},\sigma))
\ge
(2-\sqrt2)w(2R/3).
\]
Since $K-1=3$,
\[
\gPL(r^{\rm spread},\sigma)
\ge
\frac{2-\sqrt2}{3}w(2R/3).
\]
Combining the lower bound for the spread configuration with the upper bound for the clustered one---directionally valid because all curvatures here are positive---yields the ratio of Corollary~\ref{cor:separation}. Using $\sech^2x\sim4e^{-2x}$,
\[
\frac{\sech^2(R/(3\sigma))}{\sech^2(R/\sigma)}
\sim
\exp\!\left(\frac{4R}{3\sigma}\right),
\]
which gives the stated exponential growth. The two configurations share $K$ and reward range but not total Fisher trace; a trace-matched separation is not claimed here and remains open.

\paragraph{Exact illustration for group selection.}
At $R/\sigma=3$, exact enumeration of all $4!$ rankings gives $\gPL=0.0844/\sigma^2$ for $r^{\mathrm{spread}}$ and $0.0074/\sigma^2$ for $r^{\mathrm{cluster}}$, a ratio of $11.4$, with $\lambda_2(\LBT)=0.0954/\sigma^2$ and $0.0099/\sigma^2$ and Fisher traces $\tr H_\sigma=0.638/\sigma^2$ and $1.010/\sigma^2$, respectively. Choosing between the two groups by $\lambda_2(\LBT)$, as in Corollary~\ref{cor:design}, selects the spread group, which has the larger exact curvature, whereas choosing by Fisher trace selects the clustered group. This is an exact PL-model calculation, not a learning experiment.

\section{Spectral Experiment: Protocol, Validation, and Adversarial Search}
\label{app:expdetails}

\paragraph{Pre-specification.}
The protocol was frozen in a specification committed to version control before any result was generated. It was not lodged in an external registry, so we describe the experiment as pre-specified rather than preregistered; the commit history, available on request, fixes the ordering. The specification fixed the hypotheses, score families, sample counts, master seed, tolerances, figures, and interpretation rules, and was not edited afterwards. Three deviations occurred and are recorded with the experiment: a dependency version file was corrected to match the versions actually captured at runtime; one companion figure was added showing the same data with no subsetting; and a precision-stratification flag was reported alongside, never instead of, the full statistics. No hypothesis, family, sample count, tolerance, or interpretation rule was changed.

\paragraph{Instances.}
Experiment A covers $K\in\{3,4,5,6\}$, $\sigma=1$, scales $s\in\{0.25,0.5,1,2,4,8\}$, and eight families: Gaussian, uniform, evenly spaced, two-cluster, jittered two-cluster, one dominant item, one suppressed item, and a random clustered mixture, with $100$ vectors per stochastic cell and master seed $20260917$, giving $9{,}696$ instances. Experiment B is the controlled $K=4$ path at $R\in\{0.5,1,2,4,8\}$ with $t=0,0.02,\dots,1$ ($255$ points). Experiment C searches $K\in\{4,5,6\}$ at target ranges $2R\in\{1,2,4,8,16\}$.

\paragraph{Two independent implementations.}
The two code paths never call each other. The direct route computes ranking probabilities in log-sum form and forms embedded categorical covariances; the factorized route computes ranking probabilities as a forward product of stage selection probabilities, never forms a covariance matrix, and reaches the operator only through the exposure identity of Theorem~\ref{thm:factorization}. Their agreement is therefore evidence about the identity rather than a shared assumption.

\paragraph{Blocking validation.}
All ten gates passed before any scientific run: ranking probabilities sum to one ($3.2\times10^{-15}$); $H_\sigma$ symmetric (exactly); $H_\sigma$ positive semidefinite ($-5.2\times10^{-15}$); $H_\sigma\1=0$ ($7.4\times10^{-15}$); direct equals factorized ($7.5\times10^{-15}$, gate $10^{-10}$); lower sandwich ($-5.1\times10^{-15}$); upper sandwich ($-3.3\times10^{-16}$); $K=2$ Bradley--Terry equality ($1.2\times10^{-16}$); the symmetric closed form of Theorem~\ref{thm:symmetric} ($7.6\times10^{-15}$); and the $K=3$ exposure identity $\alpha_{ab}=1-q+q^2$ with $\alpha_{ab}\ge3/4$ ($6.7\times10^{-16}$, minimum $0.750141$). Over the full sweep the direct-versus-factorized relative Frobenius error never exceeded $1.03\times10^{-14}$.

\paragraph{Numerical conditioning of $\rho$.}
Thirteen of the $9{,}696$ instances produced $\rho$ outside $[1/(K-1),1]$ at double precision. Every one occurs at large score separation where both $\lambda_2$ values fall between $10^{-8}$ and $10^{-21}$, i.e. at or below the double-precision eigenvalue noise floor, so $\rho$ is a quotient of two unresolvable quantities. All thirteen were recomputed at $50$ decimal digits and all lie inside the envelope. The operator form of the theorem, which is far better conditioned, held at double precision across every instance. We therefore report the positive semidefinite comparisons as the primary validation object and treat $\rho$ as a diagnostic only.

\paragraph{Adversarial trace-matched search.}
For each $(K,\text{range})$ cell the candidate bank held $7{,}424$ vectors: the $2{,}424$ Experiment A vectors for that $K$ renormalized to the target range, plus $5{,}000$ fresh vectors drawn from Gaussian, uniform, two-cluster, three-cluster, and randomly spaced ordered families. Pairs were compared by relative trace difference and curvature ratio, scanning exhaustively within each tolerance band, so the reported value is the bank optimum rather than a selected example. Tolerances were applied in the frozen order $1\%$, $2\%$, $5\%$; the $1\%$ band already produced the separations reported in the main text and the ladder was never relaxed further. Table~\ref{tab:tracematched} gives the strongest pair at each $K$.

\begin{table}[h]
\centering
\scriptsize
\resizebox{\linewidth}{!}{
\begin{tabular}{lllrrrr}
\toprule
 & & Score vector & $\tr(H_\sigma)$ & $\gPL$ & trace gap & $\gPL$ ratio \\
\midrule
$K=4$ & A & $(-0.15692,\,-7.9407,\,0.038323,\,8.0593)$ & $0.4972135$ & $5.84\times10^{-4}$ & $0.71\%$ & $127.5\times$ \\
     & B & $(-6.2458,\,-7.9936,\,6.2331,\,8.0063)$ & $0.5007810$ & $4.58\times10^{-6}$ & & \\
\midrule
$K=5$ & A & $(0.083131,\,-0.124,\,-0.00023028,\,8.0206,\,-7.9794)$ & $1.1639773$ & $7.32\times10^{-4}$ & $0.37\%$ & $249.5\times$ \\
     & B & $(8.2913,\,-6.574,\,-6.5589,\,-4.5845,\,9.426)$ & $1.1682513$ & $2.93\times10^{-6}$ & & \\
\midrule
$K=6$ & A & $(0.17856,\,0.072203,\,-7.8465,\,-0.20526,\,8.1535,\,-0.3525)$ & $1.8811721$ & $9.83\times10^{-4}$ & $0.68\%$ & $478.3\times$ \\
     & B & $(-7.8589,\,-8.1204,\,-6.4532,\,6.7232,\,7.8796,\,7.8297)$ & $1.8939865$ & $2.06\times10^{-6}$ & & \\
\bottomrule
\end{tabular}}
\caption{Strongest trace-matched pairs at $1\%$ trace tolerance and range $16$, from the frozen adversarial search. Both members of each pair were reverified at $50$ decimal digits. This is an empirical separation over the tested candidate families, not a theorem.}
\label{tab:tracematched}
\end{table}

\begin{figure}[h]
\centering
\includegraphics[width=4.80in]{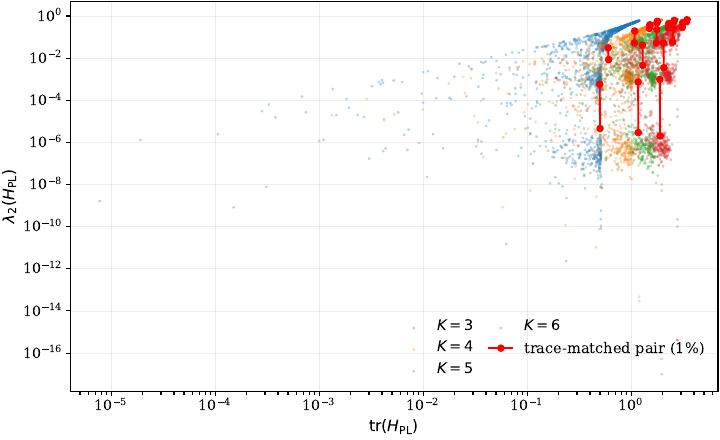}
\caption{Total Fisher mass against weakest identifiable direction over all exact instances. At a fixed trace the attainable $\lambda_2(H_\sigma)$ spans many orders of magnitude. Red segments join the trace-matched pairs located by the frozen adversarial search at $1\%$ trace tolerance.}
\label{fig:trace}
\end{figure}

\paragraph{Controlled path, full sweep.}
Table~\ref{tab:path} gives every $R$ value of the controlled deformation.

\begin{table}[h]
\centering
\small
\begin{tabular}{cccrr}
\toprule
$R$ & $\gPL$ (clustered) & $\gPL$ (spread) & $\gPL$ ratio & trace ratio \\
\midrule
0.5 & $5.16\times10^{-1}$ & $5.51\times10^{-1}$ & $1.07\times$ & $1.04\times$ \\
1   & $2.91\times10^{-1}$ & $4.00\times10^{-1}$ & $1.38\times$ & $1.11\times$ \\
2   & $5.22\times10^{-2}$ & $1.81\times10^{-1}$ & $3.46\times$ & $0.97\times$ \\
4   & $1.01\times10^{-3}$ & $4.17\times10^{-2}$ & $41.4\times$ & $0.37\times$ \\
8   & $3.38\times10^{-7}$ & $2.83\times10^{-3}$ & $8384\times$ & $0.029\times$ \\
\bottomrule
\end{tabular}
\caption{Controlled $K=4$, fixed-range path of Section~\ref{sec:controlled}, full $R$ sweep. Group size and reward range are identical at both endpoints of every row; the range equals $2R$ at every point of the path to machine precision.}
\label{tab:path}
\end{table}

\paragraph{Realized spectral ratios for the trace-matched pairs.}
For the pairs of Table~\ref{tab:tracematched}, the ratio of $\lambda_2(\LBT)$ between the two members tracks the ratio of exact curvature but is not equal to it, which is expected: Theorem~\ref{thm:sandwich} constrains each member only to within a factor $K-1$, so the two ratios may differ by up to that factor squared in the worst case. Observed differences are far smaller.

\begin{table}[h]
\centering
\small
\begin{tabular}{crrr}
\toprule
$K$ & exact $\gPL$ ratio & $\lambda_2(\LBT)$ ratio & relative difference \\
\midrule
$4$ & $127.5\times$ & $126.3\times$ & $1.0\%$ \\
$5$ & $249.5\times$ & $280.2\times$ & $12.3\%$ \\
$6$ & $478.3\times$ & $414.2\times$ & $13.4\%$ \\
\bottomrule
\end{tabular}
\caption{Exact versus reward-gap-graph curvature ratios for the strongest trace-matched pair at each $K$ (range $16$, $1\%$ trace tolerance).}
\label{tab:ratiocompare}
\end{table}

\paragraph{Why the pooled correlation is a consistency check.}
Theorem~\ref{thm:sandwich} confines $\lambda_2(H_\sigma)$ to $[\lambda_2(\LBT)/(K-1),\lambda_2(\LBT)]$. Over instances whose curvature spans many orders of magnitude, this forces a high rank and log-space correlation between the two quantities regardless of the score geometry, so a large $R^2$ cannot discriminate between hypotheses and is not independent empirical support for $\lambda_2(\LBT)$ as a predictor. What it does establish is that the implementation respects the envelope and that the envelope is tight in practice. The comparisons that carry information are with the summaries the theorem leaves unconstrained --- group size, reward range, ranking entropy, and total Fisher mass --- and with the controlled and adversarial experiments of Section~\ref{sec:experiments}, where geometry is varied deliberately rather than sampled.

\paragraph{Predictor comparison.}
Table~\ref{tab:predictors} gives the full descriptive comparison. Pooled correlations mix different $K$ and are reported for completeness only; within-$K$ values are the informative ones, and the controlled and adversarial experiments of Section~\ref{sec:experiments} are the substantive tests.

\begin{table}[h]
\centering
\small
\begin{tabular}{lrrrl}
\toprule
Predictor & Spearman & $R^2$ (log) & within-$K$ $R^2$ & Note \\
\midrule
$\lambda_2(\LBT)$ & $0.9871$ & $0.9970$ & $0.9941$--$0.9996$ & theoretically justified comparison object \\
reward range & $-0.9652$ & $0.6370$ & $0.6329$--$0.6544$ & monotone but not determining \\
ranking entropy (bits) & $0.6361$ & $0.2307$ & $0.3902$--$0.4392$ & grows with $K!$, not with $\gPL$ \\
$\tr(H_\sigma)$ & $0.6322$ & $0.2038$ & $0.3228$--$0.3639$ & Spearman $0.94$--$0.96$ within $K$ \\
$K$ & $0.1008$ & $0.0007$ & --- & pooled only; $K$ is the stratifier \\
\bottomrule
\end{tabular}
\caption{Descriptive predictor comparison for $\log\gPL$ over the $9{,}696$ exact instances. Pooled correlations mix different $K$ and are reported for completeness only; the controlled and adversarial experiments above are the substantive tests.}
\label{tab:predictors}
\end{table}

\section{Realized Exposure and Spectral Ratio}
\label{app:rho}

\begin{figure}[h]
\centering
\includegraphics[width=4.75in]{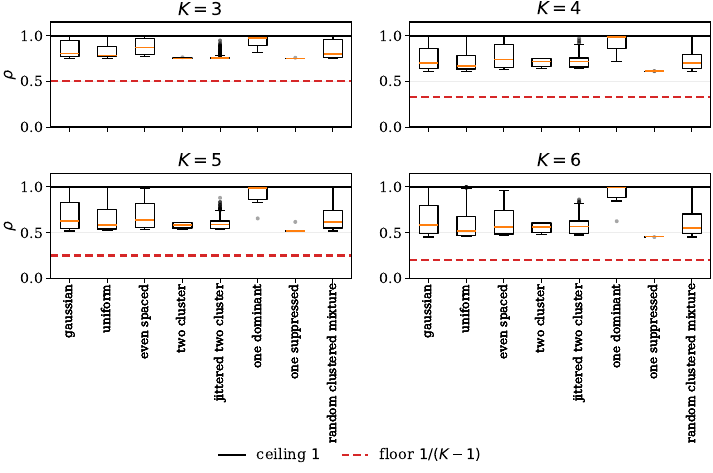}
\caption{Realized $\rho=\lambda_2(H_\sigma)/\lambda_2(\LBT)$ by $K$ and score family across the $9{,}696$ exact instances, with the theoretical floor $1/(K-1)$ and ceiling $1$. Thirteen instances for which $\rho$ is a ratio of two eigenvalues at the float64 numerical noise floor fall outside the displayed range; all were recomputed at $50$-digit precision and produced no genuine violation of the spectral bounds. They remain in the experiment and in every numerical summary.}
\label{fig:rhofamily}
\end{figure}

Every pair in every instance satisfied the exposure bounds of Theorem~\ref{thm:alphabounds}, with no violation. The smallest numerically resolvable values of $\rho$ were $0.750000$ at $K=3$, $0.610182$ at $K=4$, $0.517919$ at $K=5$, and $0.451452$ at $K=6$, against the theoretical floors $0.5$, $1/3$, $0.25$, and $0.2$. The $K=3$ minimum coincides with the exact bound $\alpha_{ab}\ge3/4$ derived in Section~\ref{sec:spectral}.

The only conclusion we draw is that the universal lower bound is \emph{conservative in the tested families}. These observations do not establish that a better universal constant exists, that $1/(K-1)$ is or is not optimal, or that symmetry is extremal in either direction; none of those questions is settled by this experiment, and Section~\ref{sec:spectral} leaves the sharpness question open.

\begin{figure}[h]
\centering
\includegraphics[width=4.75in]{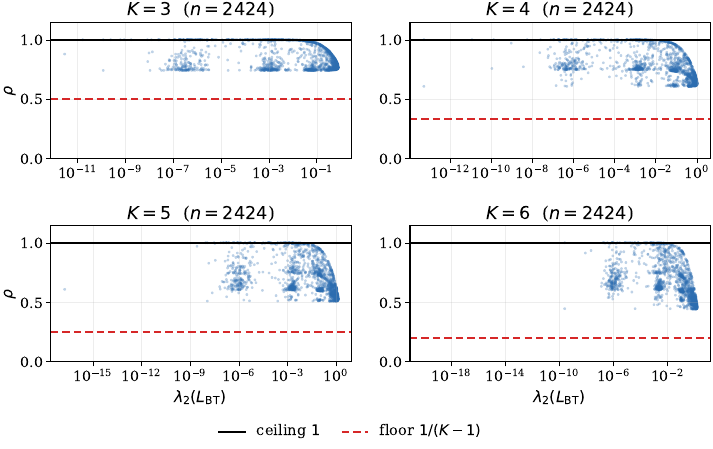}
\caption{Companion view of the data in Figure~\ref{fig:envelope}: realized $\rho$ against $\lambda_2(\LBT)$. Same instances as Figure~\ref{fig:envelope}, with none removed from the analysis; the display axis is clipped at $1.15$ to avoid numerically unresolved near-zero ratios dominating the view.}
\label{fig:envelope1b}
\end{figure}

Figure~\ref{fig:envelope1b} plots the same instances as Figure~\ref{fig:envelope}, none removed from the analysis, in a form where the width of the band is legible.

\section{Historical Supporting Experiments}
\label{app:historical}

The results in this appendix predate the spectral experiment of Section~\ref{sec:experiments} and are retained as supporting evidence. They are not the primary empirical support for any claim in this paper, and none of them is used to support a nonsymmetric curvature statement.

\subsection{Symmetric validation and the stagewise obstruction}

An earlier synthetic PL teacher experiment computed the Hessian exactly for $K\le8$ and by Monte Carlo for larger $K$. The observed nonzero symmetric eigenvalue matched, at $\sigma=1$ (the earlier code fixed $\sigma$; Appendix~\ref{app:repro}),
\[
\sigma^{-2}\frac{K-H_K}{K-1}
\]
to numerical precision in the exact regime and followed the predicted saturating curve under Monte Carlo estimation. This is superseded, for the symmetric case, by validation gate nine of the frozen experiment (Appendix~\ref{app:expdetails}), which reproduces the same closed form to $7.6\times10^{-15}$.

The same environment directly visualized Theorem~\ref{thm:obstruction}. For the explicit witness $v_S^\star$, visibility on the chosen proper subset was numerically zero, while random subsets of the same size generally retained nonzero projection energy.

\subsection{Controlled evaluator-resolution sweep}

An earlier synthetic preference-bandit experiment used a controlled noise variable $\delta$ with
\[
\sigma=\sqrt{1+\delta^2}.
\]
This mapping is an experimental mechanism, not a theoretical assumption. Table~\ref{tab:noise} reports the results.

\begin{table}[h]
\centering
\small
\begin{tabular}{ccccc}
\toprule
$\delta$ & $\sigma$ & $1/(1+\delta^2)$ & NMSE@200 & Kendall $\tau$@200 \\
\midrule
0.0 & 1.00 & 1.000 & 0.559 & 0.791 \\
0.3 & 1.04 & 0.917 & 0.560 & 0.810 \\
0.6 & 1.17 & 0.735 & 0.597 & 0.759 \\
1.0 & 1.41 & 0.500 & 0.622 & 0.739 \\
2.0 & 2.24 & 0.200 & 0.718 & 0.662 \\
4.0 & 4.12 & 0.059 & 0.869 & 0.417 \\
\bottomrule
\end{tabular}
\caption{Controlled-noise synthetic experiment from an earlier study.}
\label{tab:noise}
\end{table}

\subsection{Small real-model anchor}

An earlier study ran Pythia-410M-deduped with $K=4$, three seeds per cell, and contamination levels $\rho\in\{0,0.5\}$. The information-aware variant (IW-GRPO + step in Table~\ref{tab:pythia}) had higher mean test reward in both conditions, with paired-bootstrap differences of $+0.020$ (95\% CI $[-0.026,+0.076]$) at $\rho=0$ and $+0.013$ (CI $[-0.011,+0.027]$) at $\rho=0.5$, but lower validation exact match and validation reward than Blind GRPO at $\rho=0$. This study and the next scored candidate groups by a Fisher-trace-style utility, the first-stage softmax trace $1-\sum_a p_a^2$ of scaled rewards (optionally multiplied by a reward-variance factor), rather than by the spectral criterion of Corollary~\ref{cor:design}; these historical experiments do not test spectral group selection. In both, $\rho$ is the contamination level, not the spectral ratio of Appendix~\ref{app:rho}. This anchor is directional only and carries no headline claim in this paper; in particular it is not evidence that spectral connectivity determines language-model post-training performance.

\begin{table}[h]
\centering
\small
\resizebox{\linewidth}{!}{
\begin{tabular}{lcccccc}
\toprule
Method & $\rho$ & Val. Exact & Test Exact & Val. Reward & Test Reward & Avg. Utility \\
\midrule
Blind GRPO & 0.00 & $0.609\pm0.019$ & $0.560\pm0.030$ & $0.670\pm0.023$ & $0.616\pm0.027$ & $0.640\pm0.119$ \\
IW-GRPO + step & 0.00 & $0.551\pm0.068$ & $0.577\pm0.051$ & $0.634\pm0.038$ & $0.636\pm0.035$ & $0.644\pm0.066$ \\
Blind GRPO & 0.50 & $0.575\pm0.025$ & $0.577\pm0.042$ & $0.649\pm0.018$ & $0.633\pm0.033$ & $0.634\pm0.155$ \\
IW-GRPO + step & 0.50 & $0.623\pm0.038$ & $0.600\pm0.020$ & $0.665\pm0.034$ & $0.646\pm0.019$ & $0.748\pm0.002$ \\
\bottomrule
\end{tabular}}
\caption{Pythia-410M anchor from an earlier study; mean $\pm$ sample standard deviation over seeds $\{0,1,2\}$.}
\label{tab:pythia}
\end{table}

\subsection{Grouped post-training results}

For completeness, Table~\ref{tab:regex} reproduces the five-seed regex-lite grouped post-training results from an earlier study. These results are secondary here because they test the same trace-based selector, not the spectral criterion of Corollary~\ref{cor:design}.

\begin{table}[h]
\centering
\scriptsize
\resizebox{\linewidth}{!}{
\begin{tabular}{lcccccc}
\toprule
Method & $\rho$ & Val. Exact & Test Exact & Val. Reward & Test Reward & Avg. Utility \\
\midrule
Blind GRPO & 0.00 & $0.511\pm0.045$ & $0.524\pm0.012$ & $0.570\pm0.037$ & $0.580\pm0.009$ & $0.719\pm0.003$ \\
IW-GRPO & 0.00 & $0.489\pm0.058$ & $0.515\pm0.029$ & $0.554\pm0.044$ & $0.574\pm0.017$ & $0.740\pm0.001$ \\
IW-GRPO + step & 0.00 & $0.490\pm0.044$ & $0.536\pm0.011$ & $0.551\pm0.039$ & $0.588\pm0.010$ & $0.741\pm0.001$ \\
\midrule
Blind GRPO & 0.25 & $0.492\pm0.015$ & $0.526\pm0.018$ & $0.555\pm0.016$ & $0.583\pm0.012$ & $0.721\pm0.002$ \\
IW-GRPO & 0.25 & $0.483\pm0.031$ & $0.527\pm0.008$ & $0.552\pm0.031$ & $0.585\pm0.004$ & $0.740\pm0.001$ \\
IW-GRPO + step & 0.25 & $0.490\pm0.051$ & $0.532\pm0.006$ & $0.553\pm0.044$ & $0.587\pm0.003$ & $0.740\pm0.001$ \\
\midrule
Blind GRPO & 0.50 & $0.493\pm0.044$ & $0.513\pm0.029$ & $0.557\pm0.035$ & $0.575\pm0.024$ & $0.725\pm0.001$ \\
IW-GRPO & 0.50 & $0.498\pm0.053$ & $0.520\pm0.006$ & $0.556\pm0.040$ & $0.577\pm0.005$ & $0.741\pm0.000$ \\
IW-GRPO + step & 0.50 & $0.499\pm0.053$ & $0.532\pm0.009$ & $0.558\pm0.042$ & $0.587\pm0.009$ & $0.740\pm0.001$ \\
\bottomrule
\end{tabular}}
\caption{Five-seed regex-lite experiment from an earlier study. Paired-bootstrap test-reward differences for IW-GRPO+step minus Blind were $+0.008$ with 95\% CI $[+0.002,+0.015]$ at $\rho=0$, $+0.004$ with CI $[-0.003,+0.012]$ at $\rho=0.25$, and $+0.012$ with CI $[-0.008,+0.036]$ at $\rho=0.5$.}
\label{tab:regex}
\end{table}

\section{Reproducibility and Code Audit}
\label{app:repro}

The spectral experiment is reproducible with a single command from the frozen experiment package, which contains the specification, both implementations, the runners, the blocking validation script, the raw per-instance data, the processed summaries, the figures, and the logs. Every run records its exact command, git commit, Python and package versions, seed, timestamp, and configuration alongside its output, and no run overwrites a previous result directory.

\paragraph{Prior validation code.}
Earlier symmetric validation code used uniform ranking averaging, which is exact only at symmetry; all nonsymmetric results reported here use the independently implemented probability-weighted exact calculation. Concretely, the earlier routine averaged the $K!$ stage-Fisher matrices uniformly rather than weighting each ranking by $P_r^{(\sigma)}(\omega)$, and fixed $\sigma=1$. At exact symmetry the PL law over rankings is uniform, so the two agree there to machine precision, and the earlier symmetric validation remains valid. Away from symmetry the two differ, by up to $63\%$ in relative Frobenius norm on the 16 cases we checked ($K=2$--$5$; \texttt{spectral\_geometry/src/compare\_prior.py} and \texttt{spectral\_geometry/logs/prior\_code\_comparison.log} in the experiment package). No claim in this paper about nonsymmetric curvature derives from that earlier code.

\section{Extended Positioning}
\label{app:relatedlong}

This appendix preserves the longer discussion compressed out of Section~\ref{sec:related_work} for the main-text page budget. No citation or claim differs between the two; only the exposition is fuller here.

\paragraph{Comparison topology versus an internally generated Fisher graph.}
Vojnovi\'c, Yun, and Zhou show that optimization rates for ranking-score inference are governed by the algebraic connectivity of item-pair co-occurrence structure, with the framework covering Luce and Plackett--Luce ranking models \citep{vojnovic2023accelerated}. Seshadri, Ragain, and Ugander derive structure-dependent risk guarantees for repeated-selection models, including tight risk bounds for multinomial-logit and Plackett--Luce estimation \citep{seshadri2020rich}. Han and Xu give a unified asymptotic analysis of full, marginal, and quasi-likelihood estimators under deterministic comparison-graph and hypergraph conditions \citep{hanxu2025unified}. These works establish that spectral connectivity and comparison topology matter for PL estimation. Our object is different: the comparison design is fixed to one complete ranking, and we characterize the \emph{score-dependent Fisher graph generated internally by that ranking}. In particular, $\LBT(r,\sigma)$ in this paper is not the empirical pair-co-occurrence graph of a data-collection design; it is a reward-gap graph whose edge weights are the Fisher sensitivities of hypothetical direct Bradley--Terry comparisons.

\paragraph{Subset-wise feedback and the value of larger groups.}
Under winner-only subset feedback, Saha and Gopalan obtain PAC-ranking rates with essentially no improvement over the pairwise setting as subset size grows, whereas top-$m$ ranking feedback yields an $m$-fold improvement in their ranking objective \citep{saha2019active}. In an online subset-selection setting, full ranking feedback can yield regret improvements that scale with the played subset size \citep{saha2019relative}. These results are complementary rather than contradictory to our symmetric saturation theorem: they study repeated adaptive interaction over a universe of items, with different feedback protocols and different statistical objectives. We instead ask how much local information about the full relative-score vector is contained in \emph{one complete PL ranking}. The distinction separates the combinatorics of the observation protocol from the weakest identifiable direction of the score-space Fisher operator.

\paragraph{Optimal design and active preference acquisition.}
Thekumparampil et al. use D-optimal design to select $K$-subsets for learning a PL model under a comparison budget \citep{thekumparampil2025few}. In language-model alignment, MASS-DPO derives a PL-specific Fisher/curvature objective and uses log-determinant optimal design to select compact, complementary negative subsets within a prompt \citep{surana2026massdpo}. More broadly, Fisher- and D-optimal-design criteria are standard tools for choosing informative observations \citep{pukelsheim2006optimal}. Our goal is prior to such acquisition decisions: we characterize the exact score-space Fisher operator of a realized complete ranking. Consequently, our exposure coefficient $\alpha_{ab}$ is not a sampling probability or a learned acquisition weight; it is the exact amount by which the sequential PL ranking process attenuates the direct pairwise Fisher edge $(a,b)$.

\paragraph{Position of the present result.}
Three Laplacian objects should be kept distinct. First, classical comparison-design Laplacians encode which items co-occur across collected observations. Second, our
\[
\LBT(r,\sigma)
=
\sum_{a<b}
\Klog(r_a-r_b;\sigma)
(e_a-e_b)(e_a-e_b)^\top
\]
is a score-dependent graph of \emph{direct pairwise discriminability}. Third, $H_\sigma(r;r)=\LPL(r;r,\sigma)$ is the exact Fisher operator induced by a complete PL ranking. Theorem~\ref{thm:factorization} connects the latter two edge by edge, and Theorem~\ref{thm:sandwich} turns the exposure bounds into the uniform operator comparison of Section~\ref{sec:spectral}. Likewise, the nullspace statement behind Theorem~\ref{thm:obstruction} is elementary linear algebra; the contribution here is its structural role: it explains why positive local curvature on shrinking PL stages cannot be converted by naive scalar addition into a global bound on the weakest identifiable Fisher curvature. The fixed-range separation is a consequence of this same score-space viewpoint; the harmonic formula of Theorem~\ref{thm:symmetric} also follows from the equal-score Fisher computation of \citet{hajek2014minimax}, whose curvature comparisons use a score-independent co-occurrence Laplacian with range-dependent constants.

\section{Label Cost of the Weakest Direction}
\label{app:labelcost}

The main text characterizes $\gPL$ but does not show that it costs anything. This appendix asks whether $\gPL$ predicts estimation error, using the same controlled configurations as Section~\ref{sec:controlled}. The protocol was frozen before any result was generated and is available on request. Nothing here is a new theorem: asymptotic normality of the maximum-likelihood estimator is classical, and we are checking that the quantity characterized in this paper behaves as Fisher information should at finite $N$.

\paragraph{Design.}
For $K=4$, $\sigma=1$, $R\in\{0.5,1,2,4\}$ and $t\in\{0,\tfrac12,1\}$ along the path of Section~\ref{sec:controlled}, we draw $N\in\{100,300,1000,3000,10^4\}$ rankings exactly from the PL law by the Gumbel--max construction, fit $\hat r$ by maximum likelihood on $V_\perp$ (the PL log-likelihood is concave), and record the squared error along the eigenvector of $H_\sigma(r;r)$ for $\gPL$ and, as a control, along the eigenvector for $\lambda_{\max}$. Twelve configurations, $200$ replicates per cell. $R=8$ was excluded \emph{in advance} because its extremely small curvature, $\gPL\approx3.4\times10^{-7}$, would require millions of rankings even to accumulate order-one information in the weakest direction. Three blocking checks passed before the run: the sampler reproduces exact PL ranking probabilities (total variation $0.0015$), the estimator recovers $r$ on large samples, and $\gPL$ agrees with the exact operator.

\paragraph{The frozen hypothesis fails as stated.}
Over all $60$ cells, regressing $\log$ weak-direction MSE on $\log(N\gPL)$ gives slope $-3.750$ with $R^2=0.386$. The frozen hypothesis predicted $-1$ and is \emph{not supported}. The anomalous behaviour occurs in eight of the nine cells with $N\gPL\le6.34$ (the ninth, at $N\gPL=4.17$, has error $1.44$ times the prediction), where the estimator runs away along a direction the data barely constrains, reaching mean squared errors up to $8.8\times10^{30}$. These extreme errors are consistent with severe estimation instability in the low-information regime; Appendix~\ref{app:rebuttal:threshold} investigates this explanation and its limitations.

\paragraph{Restricted analysis (post hoc).}
Excluding cells with $N\gPL<10$, a criterion motivated by the effective sample size in the measured direction rather than fitted to the data, leaves $51$ cells and gives slope $-1.0045$ with $R^2=0.9955$, and MSE/prediction of median $1.038$ over the range $[0.72,1.28]$, with no free parameters. The control direction behaves the same way against $\lambda_{\max}$: slope $-1.0101$, $R^2=0.9975$. Within a single configuration the weak- and strong-direction errors differ by up to five orders of magnitude, so the dependence is direction-specific. This restriction was not part of the frozen specification and is reported as post hoc.

\begin{figure}[h]
\centering
\includegraphics[width=4.90in]{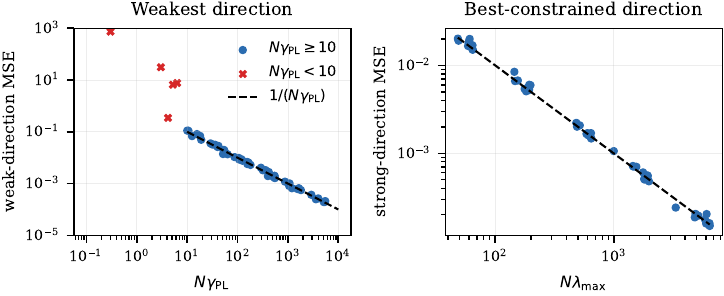}
\caption{Estimation error against effective sample size in each direction. Left: weakest direction; crosses mark cells with $N\gPL<10$, most of which exhibit severe estimation errors. Right: best-constrained direction. Dashed lines are $1/(N\gPL)$ and $1/(N\lambda_{\max})$, not fits.}
\label{fig:labelcost}
\end{figure}

\paragraph{Cost in labels.}
At matched $K$, matched range and matched $N$, the ratio of weak-direction errors tracks the inverse ratio of curvatures: at $R=2$ the curvature ratio is $3.46$ and the observed error ratios are $3.80$, $2.89$ and $3.18$ at $N=10^3,3\times10^3,10^4$; at $R=4$, where both endpoints are adequately sampled only at $N=10^4$, the curvature ratio is $41.44$ and the observed error ratio is $39.85$. The more extreme $R=4$ comparison lies at the boundary of the included regime: the clustered endpoint has $N\gPL=10.05$ at $N=10^4$. Reading this operationally, the clustered configuration needs roughly $\gamma_{\rm spread}/\gamma_{\rm cluster}$ times as many rankings to pin the weakest contrast down as well --- about $41\times$ at $R=4$ --- with the same number of candidates and the same reward range.

\paragraph{Scope.}
This establishes no finite-sample rate, no minimax statement, no claim about how groups should be \emph{selected}, and no downstream language-model consequence.

\section{Additional Robustness and Sensitivity Analyses}
\label{app:rebuttal}

This appendix collects additional robustness and sensitivity analyses. It introduces no new experiments. Section~\ref{app:rebuttal:threshold} re-analyses the frozen Appendix~\ref{app:labelcost} data. Sections~\ref{app:rebuttal:estimated} and~\ref{app:rebuttal:aggregation} are short consequences of Theorem~\ref{thm:sandwich}.

\subsection{Sensitivity of finite-sample scaling to the information threshold}
\label{app:rebuttal:threshold}

Appendix~\ref{app:labelcost} examines the empirical relationship between weak-direction estimation error and accumulated Fisher information. The regression is ordinary least squares of $\log$ weak-direction MSE on $\log(N\gPL)$, with intercept, using one observation per (configuration, $N$) cell: the mean over that cell's 200 replicates. Re-running this specification on the frozen 60-cell data reproduces the published values to machine precision:
\begin{itemize}[leftmargin=*,itemsep=1pt]
\item all 60 cells: slope $-3.74966855$, $R^2=0.38592428$;
\item the 51 cells with $N\gPL\ge10$: slope $-1.00448514$, $R^2=0.99545952$.
\end{itemize}
Table~\ref{tab:rebuttal:threshold} repeats the identical regression for a range of thresholds $t$ in $N\gPL\ge t$.

\begin{table}[h]
\centering
\caption{The Appendix~\ref{app:labelcost} regression ($\log$ weak-direction MSE on $\log(N\gPL)$, OLS with intercept) restricted to $N\gPL\ge t$, on the frozen 60 cells.}
\label{tab:rebuttal:threshold}
\begin{tabular}{rrrr}
\toprule
$t$ & cells & slope & $R^2$ \\
\midrule
0  & 60 & $-3.7497$ & $0.3859$ \\
2  & 55 & $-1.2062$ & $0.8960$ \\
5  & 53 & $-1.1420$ & $0.9127$ \\
8  & 51 & $-1.0045$ & $0.9955$ \\
10 & 51 & $-1.0045$ & $0.9955$ \\
12 & 49 & $-0.9996$ & $0.9950$ \\
15 & 48 & $-1.0033$ & $0.9949$ \\
20 & 45 & $-0.9935$ & $0.9948$ \\
30 & 44 & $-0.9922$ & $0.9945$ \\
50 & 40 & $-0.9795$ & $0.9939$ \\
\bottomrule
\end{tabular}
\end{table}

\paragraph{What the re-analysis shows.}
\begin{itemize}[leftmargin=*,itemsep=1pt]
\item \textbf{Thresholds 8--30.} The slope stays between $-1.0045$ and $-0.9922$, with $R^2\ge0.9944$.
\item \textbf{Thresholds 8 and 10} retain identical cells: no cell has $6.34<N\gPL<10.05$.
\item \textbf{Influence at threshold 10.} Leaving out any single cell gives slopes between $-1.0079$ and $-1.0006$, so no single cell drives the fit.
\item \textbf{Lower thresholds.} At thresholds 2--5 the slope is $-1.2062$ to $-1.1420$ ($R^2\approx0.90$--$0.91$). The near-$-1$ relationship therefore does not extend uniformly into the lower-information regime.
\item \textbf{Six worst excluded cells} ($N\gPL\le3.02$). We computed exactly how likely each simulated dataset is to have no finite maximum-likelihood estimate. This happens when some set of candidates is ranked above all the others in every ranking. Between about 10 and 181 of the 200 replicates per cell are expected to be like this. These are the cells whose errors are $94$ to $1.7\times10^{31}$ times the prediction.
\item \textbf{Not every anomaly is explained.} Two further excluded cells have error ratios of 34 and 49, yet expect fewer than one such replicate.
\end{itemize}

\paragraph{Limitations.}
\begin{itemize}[leftmargin=*,itemsep=1pt]
\item The threshold $N\gPL\ge10$ was chosen post hoc.
\item The frozen hypothesis over all cells was not supported.
\item This sensitivity analysis does not establish a nonasymptotic error guarantee, and the restricted relationship is not claimed to hold universally.
\item The script that generated the published regression was not preserved; the specification above was reconstructed from the text and reproduces its values exactly.
\item Per-replicate estimates and convergence diagnostics were not saved, so individual replicate failures cannot be identified.
\end{itemize}

\subsection{Robustness of spectral group selection to estimated scores}
\label{app:rebuttal:estimated}

Corollary~\ref{cor:design} selects groups by $\lambda_2(\LBT)$ computed from the true scores. In practice the scores are estimated. Let a finite candidate pool carry true scores $r$ and estimated scores $\hat r$. For a group $S$ of candidates, write $r_S$ and $\hat r_S$ for the restricted score vectors, so that $\LBT(r_S,\sigma)$ is the reward-gap Laplacian of Section~\ref{sec:setup} on $S$, written $\LBT(S,\sigma)$ in Corollary~\ref{cor:design}.

\begin{lemma}[Log-Lipschitz kernel]
\label{lem:rebuttal:kernel}
For every $\Delta\in\R$ and $\sigma>0$,
\[
\left|\frac{\partial}{\partial\Delta}\log\Klog(\Delta;\sigma)\right|
=\frac{1}{\sigma}\left|\tanh\!\left(\frac{\Delta}{2\sigma}\right)\right|
\le\frac1\sigma.
\]
Consequently $e^{-|\hat\Delta-\Delta|/\sigma}\,\Klog(\Delta;\sigma)\le\Klog(\hat\Delta;\sigma)\le e^{|\hat\Delta-\Delta|/\sigma}\,\Klog(\Delta;\sigma)$.
\end{lemma}

\begin{proof}
From $\log\Klog(\Delta;\sigma)=-\log(4\sigma^2)-2\log\cosh(\Delta/(2\sigma))$, differentiation gives $-\tanh(\Delta/(2\sigma))/\sigma$, whose absolute value is less than $1/\sigma$. The mean value theorem then bounds $|\log\Klog(\hat\Delta;\sigma)-\log\Klog(\Delta;\sigma)|$ by $|\hat\Delta-\Delta|/\sigma$.
\end{proof}

\begin{lemma}[Laplacian perturbation]
\label{lem:rebuttal:laplacian}
If $|(\hat r_a-\hat r_b)-(r_a-r_b)|\le\delta$ for all $a,b\in S$, then
\[
e^{-\delta/\sigma}\LBT(r_S,\sigma)\preceq\LBT(\hat r_S,\sigma)\preceq e^{\delta/\sigma}\LBT(r_S,\sigma).
\]
\end{lemma}

\begin{proof}
By Lemma~\ref{lem:rebuttal:kernel}, each edge weight satisfies $e^{-\delta/\sigma}\Klog(r_a-r_b;\sigma)\le\Klog(\hat r_a-\hat r_b;\sigma)\le e^{\delta/\sigma}\Klog(r_a-r_b;\sigma)$. Multiplying by the positive semidefinite $B_{ab}$ and summing over $a<b$ in $S$ preserves the Loewner order.
\end{proof}

\begin{proposition}[Group selection from estimated scores]
\label{prop:rebuttal:estimated}
Fix $K\ge2$ and $\sigma>0$. Let $\mathcal S$ be a finite family of candidate groups, each of size $K$, drawn from a pool on which
\[
|(\hat r_a-\hat r_b)-(r_a-r_b)|\le\delta
\quad\text{for all candidates }a,b\text{ in a common group of }\mathcal S.
\]
Let $\hat S$ maximize $\lambda_2(\LBT(\hat r_S,\sigma))$ over $\mathcal S$, and let $S^\star$ maximize $\gPL(r_S,\sigma)$ over $\mathcal S$. Then
\[
\gPL(r_{\hat S},\sigma)\ \ge\ \frac{e^{-2\delta/\sigma}}{K-1}\,\gPL(r_{S^\star},\sigma).
\]
\end{proposition}

\begin{proof}
For finite scores every $\Klog>0$, so each $\LBT(\cdot,\sigma)$ is the Laplacian of a connected complete graph on $K$ vertices, with kernel exactly $\mathrm{span}(\1)$. Hence $\lambda_2(A)=\min_{v\in V_\perp,\norm{v}_2=1}v^\top Av$ for every operator $A$ below. On a fixed group, by Courant--Fischer, $A\preceq B$ implies $\lambda_2(A)\le\lambda_2(B)$, and $\lambda_2(cA)=c\,\lambda_2(A)$ for $c>0$. Chaining:
\begin{align*}
\gPL(r_{\hat S},\sigma)
&\ge \tfrac1{K-1}\lambda_2(\LBT(r_{\hat S},\sigma))
&&\text{(Theorem~\ref{thm:sandwich}, lower bound, at $\hat S$)}\\
&\ge \tfrac{e^{-\delta/\sigma}}{K-1}\lambda_2(\LBT(\hat r_{\hat S},\sigma))
&&\text{(Lemma~\ref{lem:rebuttal:laplacian} at $\hat S$)}\\
&\ge \tfrac{e^{-\delta/\sigma}}{K-1}\lambda_2(\LBT(\hat r_{S^\star},\sigma))
&&\text{($\hat S$ maximizes the estimated criterion; $S^\star\in\mathcal S$)}\\
&\ge \tfrac{e^{-2\delta/\sigma}}{K-1}\lambda_2(\LBT(r_{S^\star},\sigma))
&&\text{(Lemma~\ref{lem:rebuttal:laplacian} at $S^\star$)}\\
&\ge \tfrac{e^{-2\delta/\sigma}}{K-1}\gPL(r_{S^\star},\sigma)
&&\text{(Theorem~\ref{thm:sandwich}, upper bound, at $S^\star$).}
\end{align*}
Each comparison of Laplacians is between operators on the same group and the same $\sigma$.
\end{proof}

\paragraph{Remarks.}
\begin{itemize}[leftmargin=*,itemsep=1pt]
\item \textbf{$\delta$ bounds pairwise gap errors, not score errors.} A common shift of all estimated scores does not affect it, and uniform score errors $|\hat r_a-r_a|\le\epsilon$ imply $\delta\le2\epsilon$.
\item \textbf{The bound depends on $\delta/\sigma$:} gap errors matter relative to the evaluator resolution. The same $\sigma$ is assumed in the criterion and in the truth.
\item \textbf{The two factors $e^{-\delta/\sigma}$} come from perturbing the Laplacian once at $\hat S$ and once at $S^\star$.
\item \textbf{The fixed group size $K$} gives the single constant $K-1$. More generally the lower bound holds with $K-1$ replaced by $|\hat S|-1$.
\item \textbf{$\delta=0$ recovers Corollary~\ref{cor:design}.}
\item \textbf{The guarantee is conditional.} It assumes uniformly bounded score-gap errors on the groups considered, and does not establish that practical reward models satisfy this condition.
\end{itemize}

\subsection{Aggregation across independent ranking observations}
\label{app:rebuttal:aggregation}

Let a pool of $n$ candidates carry scores $\theta\in\R^n$, and let rankings be observed independently on groups $S_1,\dots,S_m$ of sizes $K_1,\dots,K_m\ge2$. Groups may repeat or overlap. Let $E_t\in\{0,1\}^{K_t\times n}$ select the coordinates of $S_t$. Write $H_t=E_t^\top H_\sigma(\theta_{S_t};\theta_{S_t})E_t$ and $L_t=E_t^\top\LBT(\theta_{S_t},\sigma)E_t$ for the group's Fisher operator and reward-gap Laplacian embedded in the shared score space $\R^n$.

\begin{proposition}[Additive sandwich]
\label{prop:rebuttal:aggregation}
The Fisher information of the $m$ independent observations with respect to $\theta$ is $H_{\mathrm{total}}=\sum_{t=1}^m H_t$, and
\[
\sum_{t=1}^m\frac{1}{K_t-1}L_t\ \preceq\ H_{\mathrm{total}}\ \preceq\ \sum_{t=1}^m L_t .
\]
For equal group sizes $K_t=K$, this is $\frac1{K-1}\sum_t L_t\preceq H_{\mathrm{total}}\preceq\sum_t L_t$.
\end{proposition}

\begin{proof}
The log-likelihood of independent observations is the sum of their log-likelihoods. Observation $t$ depends on $\theta$ only through $E_t\theta$, so by the chain rule its Fisher information with respect to $\theta$ is $E_t^\top H_\sigma(\theta_{S_t};\theta_{S_t})E_t=H_t$, and the information of the sum is $\sum_tH_t$. Theorem~\ref{thm:sandwich} gives $\frac1{K_t-1}\LBT\preceq H_\sigma\preceq\LBT$ on each group. Congruence $A\mapsto E_t^\top AE_t$ preserves the Loewner order, and so does summation.
\end{proof}

\paragraph{Remarks.}
\begin{itemize}[leftmargin=*,itemsep=1pt]
\item \textbf{Identifiable directions.} Both bounds are positive semidefinite, so $H_{\mathrm{total}}$ and $\sum_tL_t$ have the same kernel: a positive semidefinite operator vanishes exactly where its quadratic form does, and the two bounds sandwich that form. The kernel always contains $\1_n$.
\item \textbf{Disconnected designs.} If the union of the comparison groups splits the pool into several connected components, each component's indicator vector is also unidentifiable. When the union is connected, the bounds give $\min_t\frac1{K_t-1}\,\lambda_2(\sum_tL_t)\le\lambda_2(H_{\mathrm{total}})\le\lambda_2(\sum_tL_t)$ on $V_\perp\subset\R^n$.
\item \textbf{Scope.} This is a statement about local Fisher information. It does not establish a finite-sample estimation guarantee or an optimal adaptive acquisition algorithm.
\end{itemize}

\end{document}